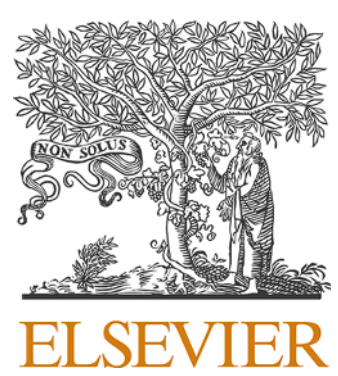

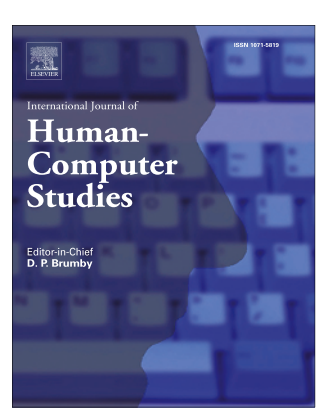

# TouchAI: Exploring human–AI perceptual alignment in touch through language model representations

Shu Zhong*, Elia Gatti, Youngjun Cho, Marianna Obrist

*Department of Computer Science, University College London, 169 Euston Rd, London, NW1, 2AE, United Kingdom*

## HIGHLIGHTS

- TouchAI reveals the gap in perception alignment between how human precieves touch experiences compared to AI.
- Human and AI's misalignment in touch is systematic and material-dependent, perceptual biases can be observed in the LLM representation space.
- Richer everyday sensory data is critical for supporting embodied human–AI alignment.

## ARTICLE INFO

## ABSTRACT

Aligning large language models (LLMs) behaviour with human intent is critical for future AI. An important yet often overlooked aspect of this alignment is the perceptual alignment. Perceptual modalities like touch are more multifaceted and nuanced compared to other sensory modalities such as vision. This study investigates how well LLMs can understand and interpret human touch experiences by focusing on their capacity to perceive the tactile qualities of everyday objects. For instance, it assesses whether LLMs can recognise that silk satin is softer and smoother than cotton denim. We developed a "Guess What Textile" interaction using a custom AI system that enables participants to narrate their touch experiences in the "textile hand" task. Participants were given two textile samples–a target and a reference–to handle. Without seeing them, participants described the differences between them to the LLM. Using these descriptions, the LLM attempted to identify the target textile by assessing similarity within its high-dimensional embedding space, where its perceptual representations are encoded. Our results suggest that a degree of perceptual alignment exists; however, it varies significantly among different textile samples. For example, LLM predictions are well aligned for silk satin, but not for cotton denim. Moreover, participants felt that their textile experiences were not closely matched by the LLM predictions. This study is the first exploration into perceptual alignment around touch using LLM encoders, exemplified through textile hand task. We discuss possible sources of this alignment variance, and how better human–AI perceptual alignment can benefit future everyday tasks.

## 1. Introduction

Human–AI alignment is a growing field that seeks to ensure artificial intelligence (AI) systems' goals, values, and behaviours are in line with human intentions (Hendrycks et al., 2020; Lee et al., 2024; Gabriel, 2020; Christian, 2020). Several studies in human–AI alignment research have discussed the imperative for AI models to "align" by having robustness, interpretability, controllability, and ethicality, to guarantee the sustainable development of AI technology (Hendrycks et al., 2020; Askell et al., 2021; Dafoe et al., 2020, 2021). These are important requirements especially as AI products and services are increasingly embedded in everyday life interactions, such as self-driving cars, smart home applications, and online shopping solutions (Elliott, 2019; ACM, 2024).

A critical but often overlooked necessity for ensuring AI models' general alignment with humans lies in their *perceptual alignment* with real world objects. Here, we define perceptual alignment as the agreement between AI assessments and human subjective judgements about the physical world, across different sensory modalities, such as vision, hearing, taste, touch, and smell. Ensuring perceptual alignment with AI is a prerequisite for achieving broader human–AI alignment and natural human–AI interactions. However, perceptual modalities vary in their explicitness and ease of evaluation (Dienes and Berry, 1997; Lynott and Connell, 2013). For instance, vision, relying on the human retina, can be effectively captured by cameras and is more straightforward to evaluate and quantify. In contrast, the sense of touch poses greater challenges, both with regard to measuring and describing touch sensations (Lynott and Connell, 2013). Unlike vision, where colours such as "yellow" can be precisely mapped to numerical values (e.g. RGB codes) and described with a shared vocabulary, touch lacks such universal standards (Hassan, 2023). Measuring tactile input depends on how force, pressure, or motion is applied and can only be partially captured through different sensor technologies (e.g. piezoresistive (Chen et al., 2019), capacitive (Mittendorfer and Cheng, 2011), or optical (Piacenza et al., 2020)). Terms like "soft" or "rough" carry no fixed numerical reference, making tactile experiences harder to quantify and label.

* Corresponding author.
*Email addresses:* shu.zhong.21@ucl.ac.uk (S. Zhong), elia.gatti@ucl.ac.uk (E. Gatti), youngjun.cho@ucl.ac.uk (Y. Cho), m.obrist@ucl.ac.uk (M. Obrist).

Recent advancements in large language models (LLMs) have showcased impressive capabilities in diverse downstream tasks and are recognised widely as foundation models (Bommasani et al., 2021). In this work, we are exploring how well LLMs can achieve perceptual alignment with humans in a textile hand task by evaluating how well LLMs can encode and represent human descriptions of textiles. In other words, we investigate how effective LLM encoders can interpret the tactile properties of textiles based on human descriptions of their touch experiences.

We focused on the concept of "textile hand"–the feel of textiles through touch (Kawabata and Niwa, 1991; Behery, 2005) - to elicit touch experiences. Tactile sensing is crucial for object discrimination (Klatzky et al., 1985; Lederman and Klatzky, 2009), and textile hand is a well-established, widely-used touch task across both academic and industrial research (Winakor et al., 1980; Atkinson et al., 2016; Petreca et al., 2013; Zhong et al., 2024). More importantly, it reflects an integral everyday task, where a good perceptual alignment would be very desirable. Imagine, for example, that ChatGPT suggests a wool sweater on a 16-degree day, which you find too warm and coarse; can it adjust its recommendation to something cooler, smoother and thinner yet adequately warm?

To this end, we designed a "Guess What Textile" interaction task and conducted an in-person user study involving 40 participants using our designed AI system. As part of the study, participants engaged in a total of 80 interaction tasks, leading to a total of 362 guessing attempts by the AI system. We analysed the AI's accuracy (i.e. its success rate in correctly guessing the textile), as well as participants' ratings of validity and similarity of AI's predictions. Our results show that while there was some degree of human–AI alignment, a bias towards certain textile samples exists (e.g. silk satin being better aligned than cotton denim). We discuss the study results in light of current human–AI alignment approaches, and highlight future research avenues considering the emergence of multimodal LLMs.

In summary, this work makes the following three contributions. We first (1) address the gap in perception alignment between human touch and AI with an in-person user study involving LLMs. To the best of our knowledge, this study represents the first exploration of the level of alignment between human touch experiences and LLMs. Our approach emphasises the importance of using comparative measures, such as validity and similarity, which encompass human subjective judgment in an interactive experience. We then (2) map sensory information onto the embedding space and measure representation similarities to encode semantics effectively in LLMs. Our observations indicate that LLMs exhibit perceptual biases across various textiles—showing significantly greater alignment with human perception for certain textiles compared to others. Finally (3) we present the design of an interactive task to study human–AI alignment, mirroring a real-life scenario. While we focused on the sensory experience of textiles, this interactive task can also be used in other everyday sensory interactions, such as selection of foods (e.g. choice between sweet fruits) or perfumes (e.g. recommendation of floral fragrances). Aligning human–AI sensory perception can have a profound impact on people's daily lives in the future.

## 2. Related work

In this section, we review relevant prior works on human–AI alignment, specifically perceptual alignment with a focus on the human sense of touch.

### 2.1. Human–AI alignment

Human–AI Alignment refers to the design, development, and refinement of artificial intelligence systems that understand, predict, and augment human intentions and behaviours (Hendrycks et al., 2020; Sucholutsky et al., 2023). Hendrycks et al. (2020) presented the ETHICS dataset to evaluate language models' understanding of values—basic moral principles, such as justice, well-being, and commonsense morality. Further research has built upon this work to reduce toxicity and promote ethical behaviour in language models with human-in-the-loop (Ouyang et al., 2022; Goyal et al., 2022; Achiam et al., 2023; Pan et al., 2023).

There is growing research into representational alignment between AI and humans. Representational alignment refers to the extent to which the internal representations of two or more information processing systems are aligned (Sucholutsky et al., 2023). This idea has been explored under various terminologies across different contexts, such as latent space alignment (Tucker et al., 2022), conceptual alignment (Stolk et al., 2016), systems alignment (Aho et al., 2022), representational similarity (Kriegeskorte et al., 2008; Muttenthaler et al., 2022), model alignment (Moayeri et al., 2023), and representational alignment (Sucholutsky et al., 2023). Christian (Christian, 2020) provides a narrative exploration of the challenges and opportunities in human–AI interaction, emphasising the importance of designing AI systems that understand and adapt to human emotions and social norms.

Extending the exploration of Human–AI alignment into sensory judgements, Marjieh et al. (2023) demonstrated that GPT-4 can effectively interpret certain human sensory judgements (e.g. colour, sound and taste) based on textual sensory inputs. For example, they displayed the same pair of colours (red and blue) to both humans and GPT models,[1] requesting each to rate the similarity score, and then comparing the resulting scores. Their findings show that judgements made by GPT models exhibit correlations with those made by humans. Zhong et al. (2024) investigates how effectively Multimodal Large Language Models (MLLMs) can understand and describe the tactile sensations of textiles.

Building on prior works, we aim to understand how current AI models interpret and comprehend semantic sensory experiences, measuring the agreement between humans and AI.

Our research extends these exploratory text-in-text-out studies with LLM decoders by investigating LLM's representation (Bengio et al., 2013) in the latent space with encoders. LLMs primarily process and generate text-based information using language, and they do not inherently understand or interpret vision, hearing, touch or other sensory modalities. These models process data to create high-level embeddings that provide the model with crucial information about the data's key characteristics—they represent concepts and ideas within a latent space or embedding space (Bengio et al., 2013). In embedding spaces, embeddings are learned representations, such embeddings enable the grouping of semantically similar items and keep dissimilar items apart in the embedding space (Hinton and Roweis, 2002). The embedding model plays a key role in the LLM architecture, and examining the encoder is essential for understanding how the model processes and arranges information—both in terms of dimensional and categorical structure (Bricken et al., 2023). Gaining insights into the encoder's function is important, as it reveals how the LLM organises its internal knowledge (Elhage et al., 2022). Similar to Zhong et al. (2024), we employ an LLM encoder to explore whether the model has incorporated tactile information into its representations.

[1] Given that GPT models lack the ability to "see" colour hex codes are provided as textual inputs.

### 2.2. AI touch perception alignment

Unlike vision, where colours can be mapped to precise numerical codes, touch has no universal standard (Hassan, 2023). Tactile sensing captures surface properties in various ways, such as piezoresistive (Chen et al., 2019), capacitive (Mittendorfer and Cheng, 2011) and vision-based sensors (Agarwal and Man, 2021). Touch has long posed challenges in HCI and robotics, where tactile understanding is critical for manipulation, human–robot interaction, and assistive technologies. Yet the field still lacks universally recognised metrics for evaluating tactile alignment with diverse sensors used in touch-related tasks.

People use natural language to communicate how things feel to the touch. We share a broadly common understanding of tactile qualities: glass is hard and smooth; silk is soft; cement is rough. What varies is the degree or dimension of emphasis. One person may describe cement as very rough, another simply as rough. Touch perception is multidimensional, typically described along dimensions such as soft–hard or rough–smooth (Okamoto et al., 2012). Prior HCI work has examined how people verbalise touch. For example, Obrist et al. (2013) showed that people often use different and sometimes inconsistent descriptors. MacLean et al. (2017) noted the lack of standardised vocabularies across contexts in multisensory design. Yet such differences reflect shifts in linguistic expression or in which tactile qualities people find most noticeable, rather than disagreement over how things feel. Cross-cultural research also supports this shared perceptual understanding: Okamoto et al. (2012) found strong cross-language agreement in fundamental haptic dimensions such as roughness and hardness. These nuances show that while language can express much about how things feel, it can never capture the full complexity of touch. In this work, we examine whether LLMs encode similar shared tactile knowledge(e.g. silk is softer than cotton denim), or whether their representations diverge from human judgements.

Such variation does not mean people fundamentally disagree about tactile reality, for example, cement and concrete are always rough and hard, but they differ in degree (e.g. very rough versus slightly rough). These graded differences highlight both the importance and the limitations of language as a medium for communicating tactile perception. Building on this prior work, we ask whether current LLMs also encode tactile attributes in ways that align with human perception, or whether their interpretations diverge.

Recent work has begun to address tactile alignment in multimodal AI. Fu et al. (2021) introduced a Touch-Vision-Language (TVL) dataset pairing tactile sensor images with visual data, annotated with a mix of human labels (10%) and model-generated pseudo-labels (90%). For instance, a sample may include the tactile scan of a cotton fabric, a photo of the garment, and a label such as “coarse, woven, deformable”. Using this dataset, they trained a vision-language-aligned tactile encoder and a TVL model, reporting a top-1 tactile-vision accuracy of 79.5% but much lower performance for tactile-language (36.7%) and vision-text (28.4%). These results suggest that vision dominates the multimodal accuracy, and tactile-language mapping remains in its early stages. Our study complements this work by evaluating how widely used foundation models (e.g. OpenAI family) represent human tactile descriptions in an interactive textile recognition task. While Fu et al. (2021) advance dataset-driven multimodal alignment and domain-specific models, we assess the alignment capacity of general-purpose models already integrated into HCI practice.

While we see advances in human–AI alignment, they are still mostly limited to vision, exemplified by recent efforts from Boggust et al. (2022); Marjieh et al. (2023) and Lee et al. (2024). These human–AI alignment measurements are often based on inter-rater reliability (IRR) to measure model and human alignment. While IRRs can measure agreement levels, they do not adequately capture deeper nuances in perceptual alignment, such as the subjective assessment of softness. For example, if someone prefers the touch of a material slightly softer than cotton denim, will the AI be able to judge this as cotton corduroy or silk satin? When it comes to our sense of touch, it is important to account for those subjective assessments to ensure a satisfying human–AI interaction.

Research in haptics has also sought to capture perceptual similarity between materials. Pairwise ratings (Vardar et al., 2019; Richardson et al., 2022), cluster-sorted multidimensional scaling (Pasquero et al., 2006), and simulation-guided aggregation techniques (Lim et al., 2025) explore how humans structure texture spaces. Other work has proposed constructing haptic attribute spaces with dimensions such as rough–smooth or sticky–slippery, Petreca et al. (2013); Guest and Spence (2003) sometimes predicted directly from images (Hassan, 2023). These approaches demonstrate that while language provides one route for describing touch, perceptual alignment can also be studied through structured similarity metrics. In our study, we investigate whether the textures recognised as similar by humans are also perceived as similar by LLMs.

### 2.3. Textile hand: a concept to evaluate human-AI perceptual alignment

Textile hand, also known as fabric hand or the “hand” of a textile, refers to the feel, texture, and overall tactile qualities of a textile when touched against the skin (Kawabata and Niwa, 1991; Behery, 2005). This task encompasses a broad spectrum of tactile attributes and permits comprehensive verbalisation of touch experiences. Characteristics such as softness, smoothness, warmth, elasticity, and thickness define a textile’s hand (Ciesielska-Wrobel et al., 2012). The hand of a textile is influenced by the materials used, the weave or knit structure, and the finishing processes applied (Choi and Ashdown, 2000). In fashion and interior design, the hand of a textile is critical because it influences the textile’s look, functionality, and comfort (Slater, 1977). Designers and manufacturers prioritise the understanding of textile hand to select the best materials for their products and create satisfying subjective experiences (Petreca et al., 2015; Petreca, 2016), with growing interest in leveraging technology to recognise textile properties and consumer preferences through touch exploration (Kitagishi et al., 2023; Olugbade et al., 2023).

Moreover, as suggested by Guest and Spence (2003), real textile samples provide benefits such as precise control over textural properties, everyday relevance, and being safe for repeated exploration. Textiles allow systematic control of textural variation: textural properties can be engineered through measurable parameters such as fibre type, yarn density, weave structure, or finishing treatments, enabling comparisons of specific tactile attributes (e.g. softness, smoothness, stiffness). They also provide everyday relevance, as they are central to people’s daily tactile experiences (e.g. clothing, bedding, furniture). Finally, they are also gentle and non-abrasive to the skin, supporting safe and repeated exploration without discomfort, unlike sharp or abrasive surfaces. These characteristics make textiles widely used as a testbed for tactile evaluation and motivate the choice of textile hand in this study.

## 3. System design

We designed an AI system to facilitate “Guess What Textile?”, an interactive task in which users describe their tactile experiences with various textiles, allowing the AI to guess the textile.

In this section, we describe the overall design of “Guess What Textile“ task and the AI system, later used in our user study. We explain the selection of the textile samples based on a combination of expert descriptions and a domain-focused taxonomy (Zhong et al., 2023). This selection process resulted in 20 LLM-generated vectors $\mathcal{E}_{textiles} = \{v_1, v_2, \dots v_{20}\}$, where $v_i$ represents an embedding vector generated by an LLM. We then describe how we developed an AI prediction flow in Section 3.4 using these pre-built embeddings $\mathcal{E}$.

**Fig. 1.** The overall design of the "Guess What Textile?" task. Participants touch two textiles (a target and a reference textile) placed inside a box to hide any visual influences. The AI guessing system knows only the reference textile and is required to make a prediction of the target textile based on participants' descriptions. The task is iterative, and stops only when a correct prediction is made or when the maximum number of five attempts is reached. ASR stands for Automated Speech Recognition.

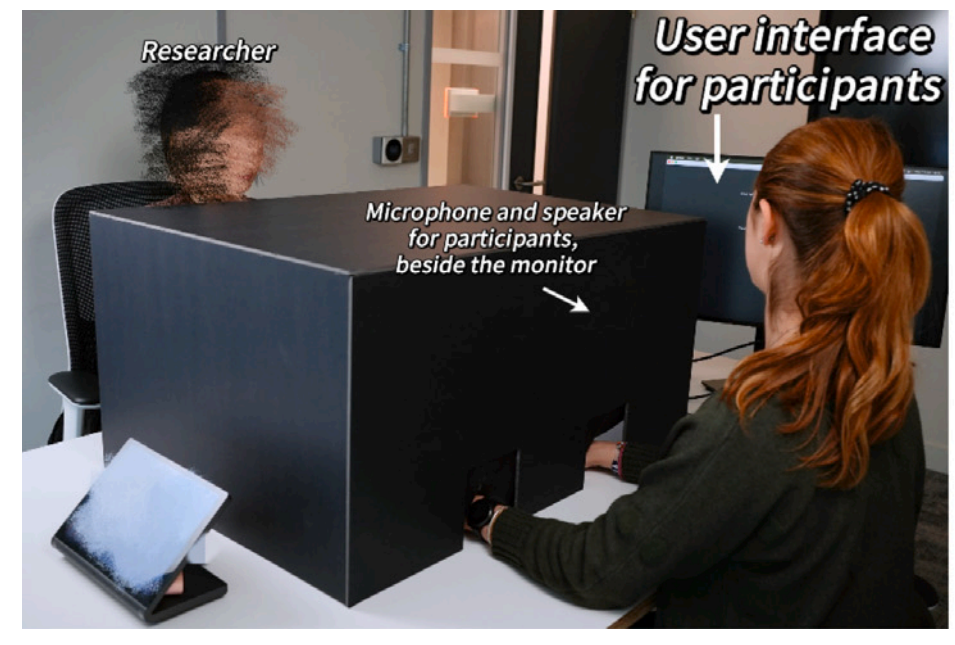

(a) A participant setup for the user study.

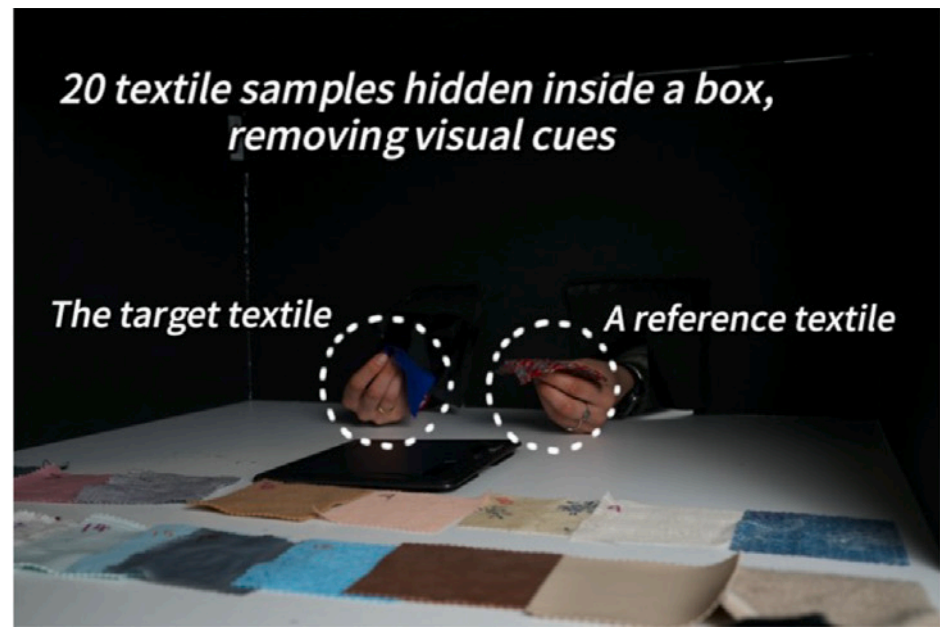

(b) A participant exploring textiles with their hands.

**Fig. 2.** Overview of the user study setup: (a) A participant sitting comfortably at a desk and putting their hands through dedicated openings in a black box that contains the textile samples. Opposite the participant, the researcher provides instructions and provides participants with the textiles selected by the AI system as part of the study task. Each textile sample has a unique ID. (b) A participant handling one textile sample in each hand, one representing the reference textile (starting point for the "Guess What Textile" task) and the target textile described to the AI system.

### 3.1. *"Guess what textile" task design*

We selected the textile hand concept to investigate touch experience for several reasons. First, it is an everyday interaction, that people are familiar with, starting from the choice of clothes to wear every morning or running a hand over. Second, textile hand is a well-established, widely-used touch task across both academic and industrial research (Winakor et al., 1980; Atkinson et al., 2016; Petreca et al., 2013). Textiles combine everyday relevance with controlled variation of tactile properties, while remaining safe for repeated exploration, making them suitable for user studies (Guest and Spence, 2003). Lastly, textile descriptions are widely used in online fashion retail websites, catalogues and books, serving as training data for web-scale models like GPTs (Radford et al., 2019).

We developed an interactive system that enables participants to engage with AI by experiencing textiles through touch. The system is designed as a *comparative description task* that supports *continuous Human–AI interaction,* as shown in Fig. 1. In the "Guess What Textiles?" task, participants provide verbal descriptors of their sensory experience while handling two textiles at a time (i.e. a target and a reference textile sample). This name is inspired by the descriptive guessing game *"Guess Who?"*.[2] In our study, the textiles are placed inside a black box so that participants can only touch, not see them, preventing visual influence as shown in Fig. 2.

The AI system knows the reference textile's ID and descriptions of all textiles. The AI system's aim is to identify the target textile accurately based on the participant's descriptions. If the system makes an incorrect prediction, the predicted textile becomes the new reference. The participant is then asked to compare and describe the touch sensation using this updated pair. The interactive task continues until the correct prediction is made or when a maximum of five predictions is reached. This attempt limit was introduced for study purposes, ensuring consistency and comparability across participants (see more details in the Study Design in Section 4).

### 3.2. *Textile sample selection*

We selected 20 textile samples based on the TextileNet taxonomy (Zhong et al., 2023). This taxonomy has four meta-categories (i.e. fibre types): natural, animal, regenerated, and synthetic fibres. Two popular materials from each category were chosen based on annual consumption data (Coppola, 2021). Each sample was sourced from commercial sample books (Texflag Sample Book Co, 2023)–a tool used by fashion and furniture design professionals–and consists of only a single fibre type (100%). We chose textiles produced through different methods (e.g. silk crepon and silk satin) from a pool of over 100 samples, ensuring a variety in texture and physical properties. Fig. 3 shows an overview of the 20 selected textile samples used in the user study.

The sample descriptions were derived from Textilepedia (Chan, 2020) and the providers' sample books (Texflag Sample Book Co, 2023), detailing composition, fabric names, production characteristics, and common uses. The description template is organized as follows: *[Textile Name] is [characteristic]. [Additional information from sample book]. [Composition], is a [fibre category type] produced by [raw material], [characteristic of fibre]. [Fabric] is [production method] and [characteristic]. [Textile name] is commonly used for [application].*

[2] "Guess Who?" is a two-player guessing game where players deduce each other's chosen character by asking about physical traits like hair colour and presence of eyes.

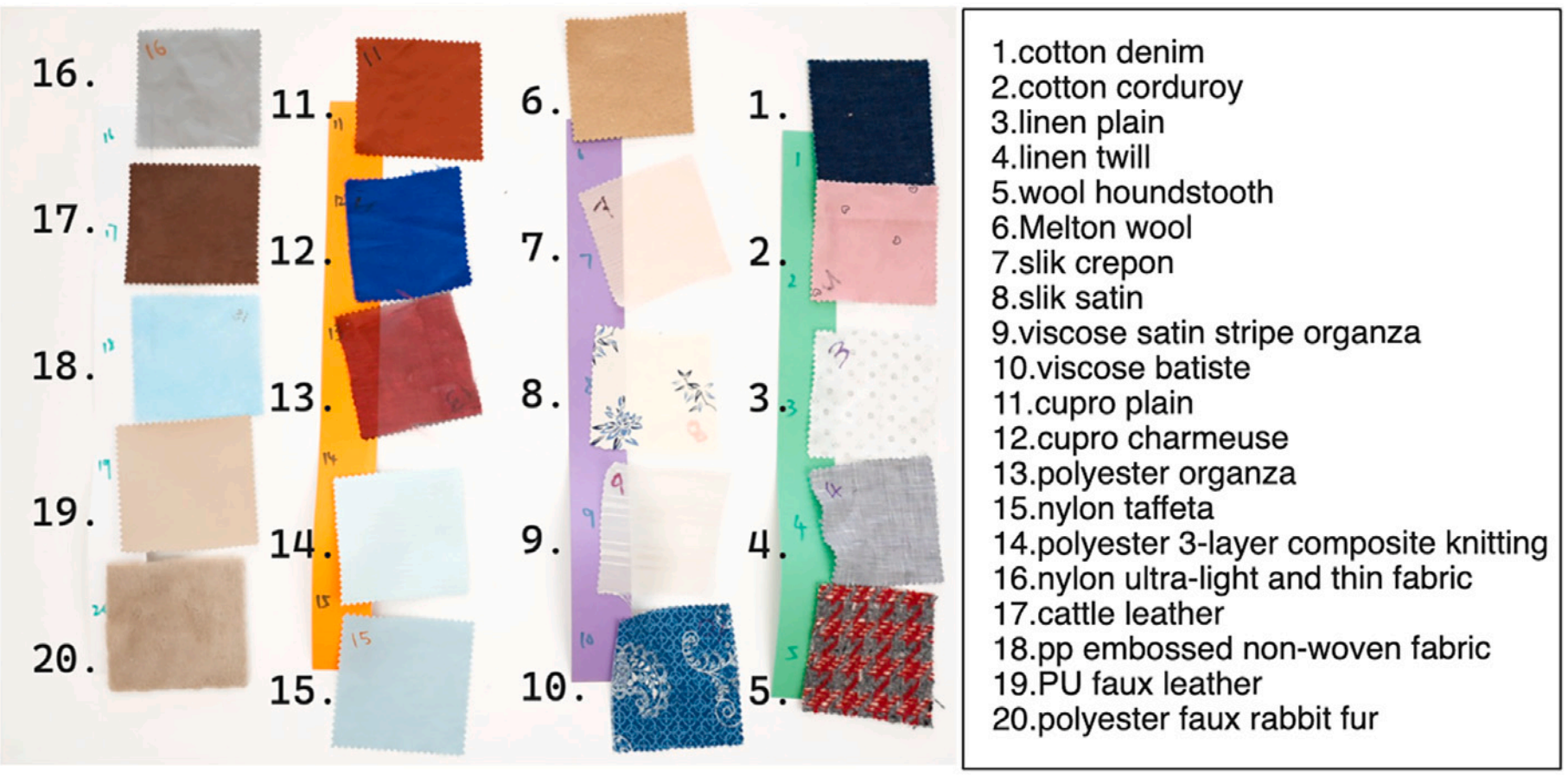

**Fig. 3.** Overview of the 20 textile samples selected for the user study.

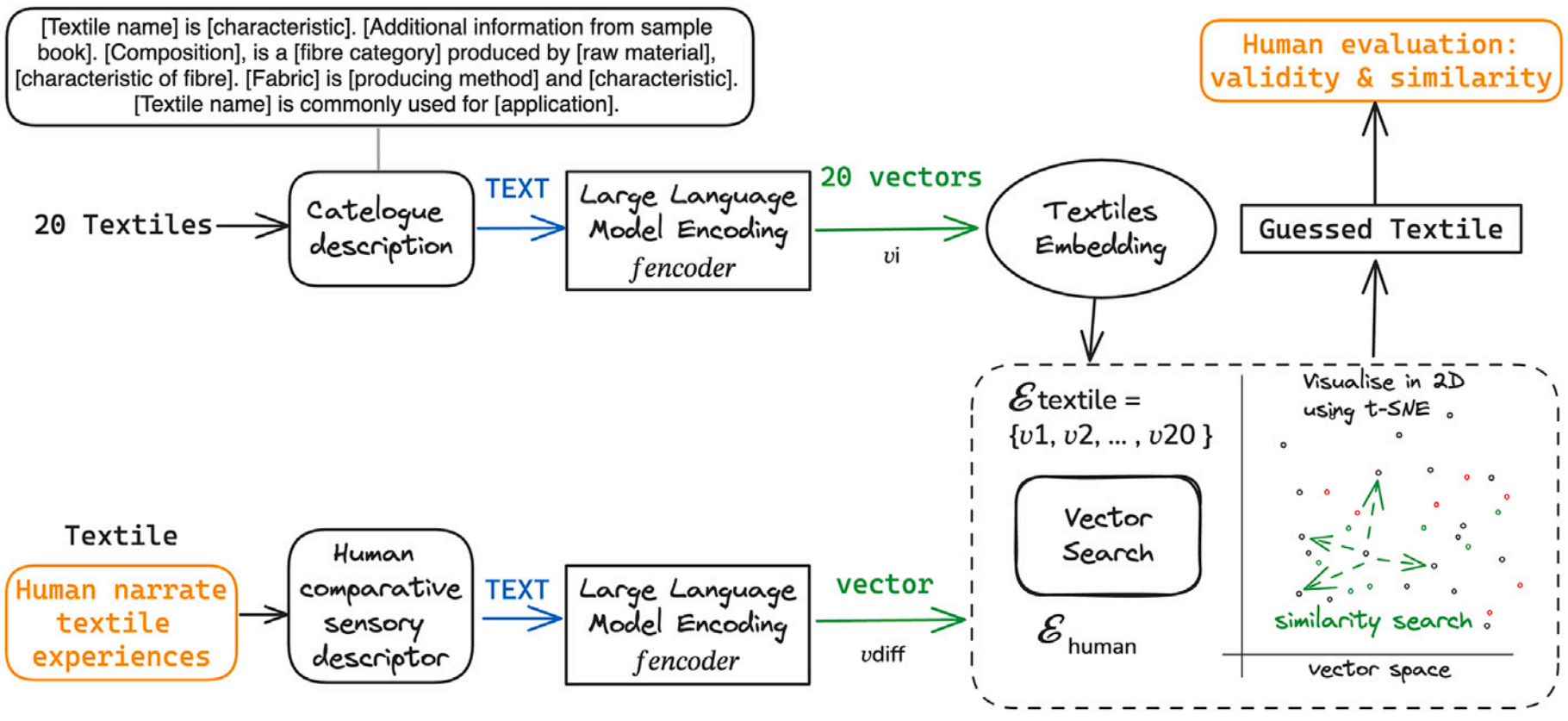

**Fig. 4.** An overview of the AI Guessing System, i.e. “Guess What Textile?”. The vector search process uses pre-built embeddings for 20 textile samples and compares them with a user query-generated vector to identify the best matching textile ID.

Both the sample selection process and the descriptor template generation are carefully discussed and consulted with domain experts, including designers and material scientists. This ensures that our descriptions follow common practices in the textile sector and cover a broad spectrum of textile properties in our sample pool. We provide a full list of descriptions in our Appendix.

### 3.3. Textile description generation

As part of our study, we examine the LLM’s behaviours by analysing their embeddings. Hence, we use model embeddings as part of our AI system design. Model embeddings, generally derived from the “internal states” of a pre-trained model, offer a way to gauge semantic similarity in the vector space. Mathematically, an embedding is a learned representation of inputs, designed to capture the semantics of the input data and make it easier to perform other machine learning tasks (Hinton and Roweis, 2002). This enables the proximity mapping of semantically ambiguous entities or textual elements to be located close to specific concepts. For instance, a cat and a dog are placed closer in the embedding space than a cat and a banana. In our particular case, similar textiles, based on their descriptors, should cluster in proximity. In other words, we examine whether LLMs grasp the concept of a “softer textile” by checking whether such textiles indeed cluster closely in the LLM’s embedding space.

To enable real-time interaction, we have developed an AI guessing system (Fig. 1), and the details of this system are further illustrated in Fig. 4. For the AI guessing system to make a prediction of the target textile sample, it first needs to encode the 20 possible textiles into LLM embeddings (top row in Fig. 4). We employ OpenAI’s `text-embedding-3-small` (Neelakantan et al., 2022) to create our embeddings, as this model is among the top-performing ones on the market.

Each of the 20 textile sample descriptions, $x_i$, was encoded into a fixed-length vector using the embedding model, $f_{encoder}$ defined as:

$$v_i = f_{encoder}(x_i) \tag{1}$$

This produced the pre-encoded textile set $\mathcal{E}_{textiles} = \{v_1, v_2, \ldots, v_{20}\}$, where each vector represents a different textile. These vectors form the AI system’s “knowledge base” for performing vector search during the user study. Unless otherwise specified, all references to “the embedding model“ in this section refer to this LLM embedding model `text-embedding-3-small`.

### 3.4. AI guessing system

The function of the AI guessing system is to capture participants’ verbal descriptions, map them into a shared embedding space, and identify the closest match from a set of pre-encoded textile embeddings as shown in Figs. 4 and 5.

The system includes several components. A graphical user interface (Fig. 6) supports interaction and feedback during the study. During each trial, participants provide comparative descriptions of a target textile relative to a reference textile. These descriptions are encoded into vectors by the same embedding model used to construct $\mathcal{E}_{textiles}$. The AI mechanism then compares input vectors with pre-built embeddings and selects the most similar candidate. The system also records participants' responses and interaction histories. The process follows standard embedding and vector search methods widely used in natural language processing and information retrieval (Hinton and Roweis, 2002; Bricken et al., 2023; Mikolov et al., 2013; Reif et al., 2019). The pipeline consists of three main steps: mapping human touch experiences into the embedding space, prediction via similarity search, and iterative comparison if the AI predicts incorrectly.

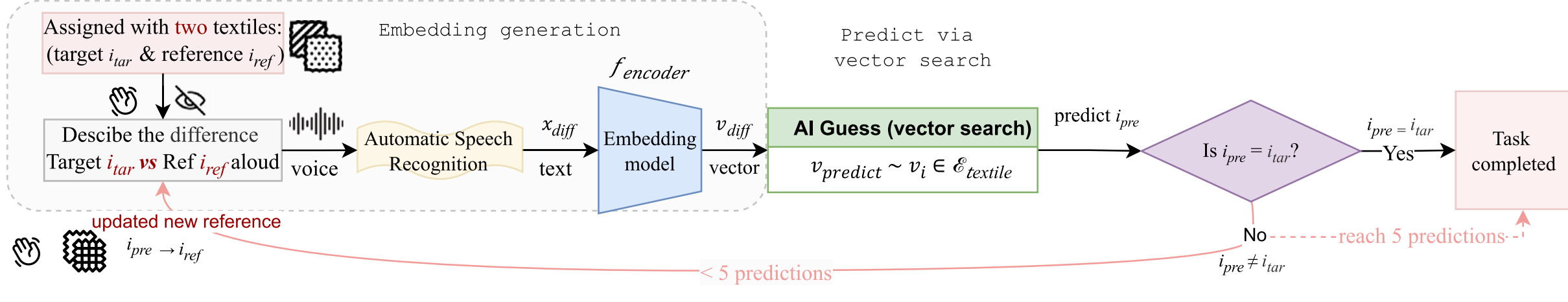

**Fig. 5.** Task algorithm overview. Participants are assigned two textiles (target and reference) and describe their tactile differences aloud. Speech input is transcribed into text and encoded as an embedding vector. The system predicts the target textile by comparing this query embedding with precomputed textile embeddings, $\mathcal{E}_{textiles}$, through vector search. If incorrect, the predicted textile becomes the new reference, and participants provide new comparative descriptions. The process repeats until the system makes a correct prediction or five attempts are reached.

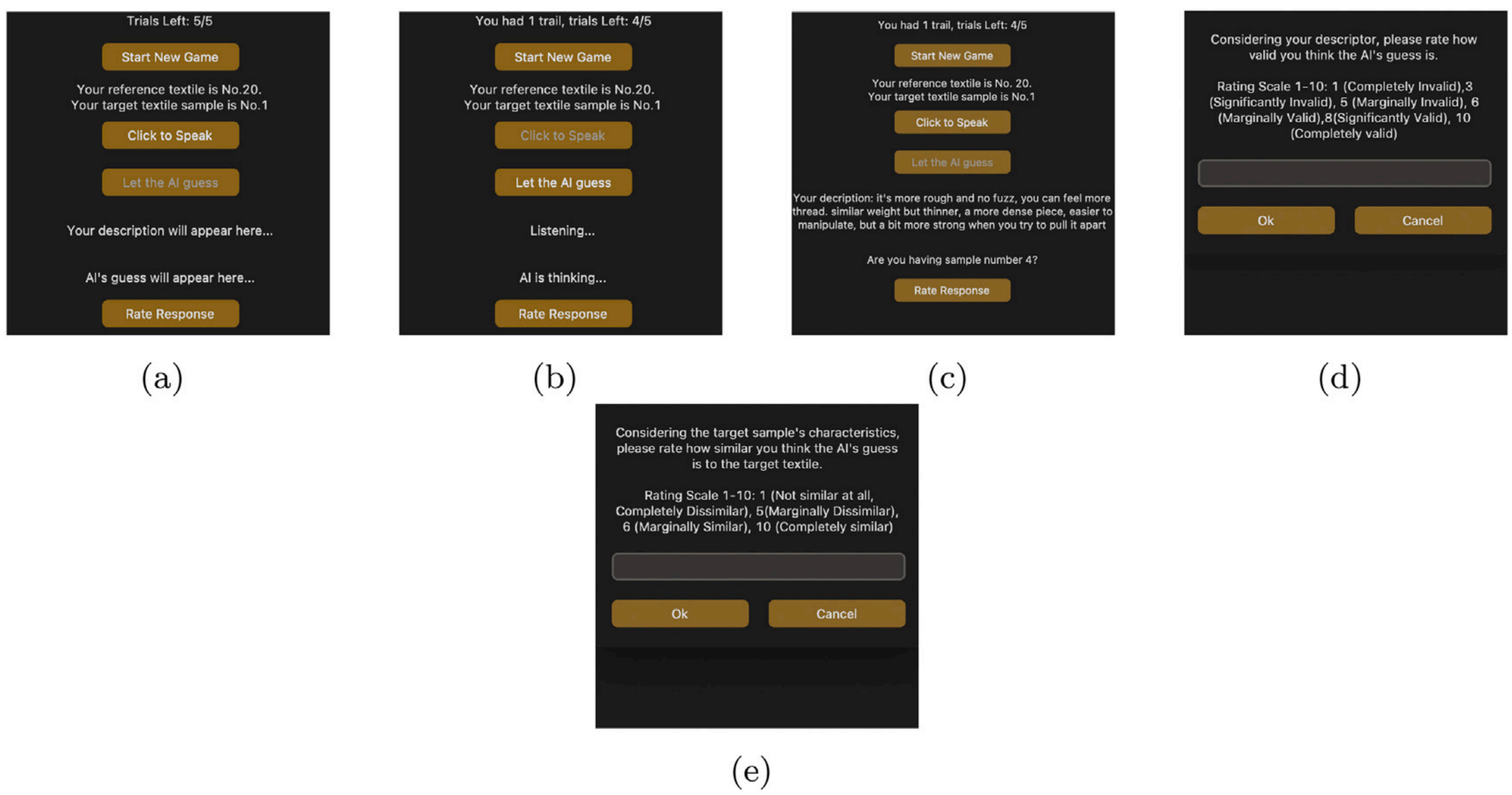

**Fig. 6.** Screenshots of the user interface used in the user study. (a) The AI system assigns two textile samples to a participant; (b) Participants describes their textile experience of the target textile; (c) the system makes a prediction for the target textile based on participant's descriptions; (d) participant rates the AI's response validity based on a scale from 1 to 10 orally; (e) participant provides a similarity score on a scale from 1 to 10 orally, comparing the AI-guessed textile to the actual target textile. Participants' responses are captured and entered by the researcher sitting opposite the participant as shown in Fig. 2.

### 3.4.1. Similarity scores

Participants also rated the perceptual similarity between the AI's guess and the target textile on a 1–10 scale (1 = not similar at all, 10 = completely similar). Recognising that no two textiles in our sample pool are identical, this similarity score focuses on the degree to which the textiles are alike in perceived characteristics. Even when the AI's prediction was incorrect, a high similarity score indicated that it still captured relevant tactile attributes such as roughness.

*Mapping human touch experience descriptions into the embedding space.* Verbal descriptions are encoded into vectors using the embedding model. When a participant provides a comparative description, the system encodes it to produce $v_{diff}$ which can then be directly compared with the pre-encoded textile embeddings $\mathcal{E}$:

$$v_{diff} = f_{encoder}(x_{diff}) \tag{2}$$

where $x_{diff}$ is the participant's comparative description and $f_{encoder}$ denotes the LLM embedding function.

*Prediction mechanism.* The system begins with the embedding of the reference textile, $v_{ref} \in \mathcal{E}_{textiles}$. Once a participant's comparative description has been encoded, the system combines $v_{ref}$ with $v_{diff}$ to produce a prediction vector. The prediction embedding is updated by first applying a directional offset and then normalising the result:

$$\tilde{v}_{predict} = v_{\text{ref}} + \alpha \frac{v_{diff}}{\|v_{diff}\|} \tag{3}$$

$$v_{predict} = \frac{\tilde{v}_{predict}}{\|\tilde{v}_{predict}\|} \tag{4}$$

where $\|v\|$ is the Euclidean norm of a vector $v$, calculated as $\sqrt{vv^T}$, and $\alpha$ is a scaling factor controlling the influence of the comparative description. We set $\alpha = 2$ for all experiments, selected via a pilot ablation study to balance the contributions of the comparative description and the reference embedding. The prediction vector $v_{predict}$ is then compared against the each textile in $\mathcal{E}_{textiles}$ using cosine similarity:

$$\cos(v_{predict}, v_i) = \frac{v_{predict} \cdot v_i}{\|v_{predict}\| \, \|v_i\|}, \quad v_i \in \mathcal{E}_{textiles} \tag{5}$$

The system selects the most similar textile by taking the maximum similarity:

$$ID = \arg\max_i \cos(v_{predict}, v_i), \quad i \in \{1, \ldots, 20\} \tag{6}$$

Here, $ID$ denotes the index of the textile with the highest similarity score, i.e. the system's best guess of the target textile.

*Iterative comparison.* If the prediction is incorrect, the guessed textile becomes the new reference. The participant then provides another comparative description of the target relative to this updated reference. Each new description $x^{(t)}_{diff}$ is encoded into a vector $v^{(t)}_{diff}$, and the prediction vector is updated recursively:

$$\tilde{v}^{(t)}_{predict} = v^{(t-1)}_{predict} + \alpha \frac{v^{(t)}_{diff}}{\|v^{(t)}_{diff}\|} \tag{7}$$

$$v^{(t)}_{predict} = \frac{\tilde{v}^{(t)}_{predict}}{\|\tilde{v}^{(t)}_{predict}\|} \tag{8}$$

where $t$ denotes the trial index ($t = 1, \ldots, T$), and $T$ is the maximum number of attempts (set to 5 in our study, see Section 4.1.1). The update accumulates comparative cues across trials, allowing the system to refine its prediction with each additional description.

This iterative loop continues until the system makes a correct prediction or a maximum of five attempts is reached. With each additional description, the query vector incorporates more information, increasing the likelihood of convergence on the correct textile. In theory, with more information provided, the AI is more likely to predict correctly if the LLM's embedding space is well aligned with human tactile semantics.

## 4. User study design

We designed a user study to elicit human touch experience descriptors using the "Guess What Textile?" system described in the previous section. The aim of the study was to understand how well LLMs can interpret and predict the textile a person is handling based on their touch experience descriptions—the alignment in touch-related representations. The study was conducted between February and March 2024, and was approved by the local University Research Ethics Committee.

### 4.1. Measuring human-AI alignment

We measure the degree of perceptual alignment between humans and LLMs using three evaluation metrics:

- AI Success Rate: whether the AI correctly identified the target textile within a limited number of guesses.
- Validity: how well the AI's prediction matches participants' verbal descriptions.
- Similarity: how close the guessed textile feels to the actual target.

In addition to our main evaluation metrics, we conducted an exploratory linguistic analysis to examine whether individual differences in participants' verbal descriptions contributed to variation in AI performance across textiles. We computed participants' description similarity using a lexical approach (TF-IDF), a semantic embedding approach (Sentence-BERT) and contextual understanding (`text-embedding-3-small`).

#### 4.1.1. AI success rate

Each task allowed the AI up to five attempts to identify the target textile, with the process stopping early if the correct textile was predicted. This five-attempt limit was determined through a pilot test with six participants, where we observed that in cases where the AI repeatedly predicted the wrong textile, participants might run out of descriptors and lose confidence in continuing. Ideally, the AI system could continue guessing until it identified the correct textile, but allowing unlimited trials would have made interactions overly long and reduced response quality. The five-attempt cap therefore balanced experimental consistency with participant comfort.

A task was considered *successful* if the AI identified the correct target within these attempts. We refer to this measure as a *success rate* rather than accuracy, as it reflects whether the AI succeeded within a limited number of interactive trials.

We report two levels of success rate: (1) the overall success rate, calculated as the proportion of correctly identified textiles across all completed tasks, and (2) the textile-specific success rate, examining whether alignment varies across material categories (e.g. silk, denim).

Because objective success alone cannot fully capture perceptual alignment, we complemented it with two subjective measures, *Validity* and *Similarity*, which participants rated only when the AI's prediction was incorrect. These measures help distinguish whether misalignment arose from language interpretation or from limitations in the model's tactile understanding.

#### 4.1.2. Validity scores

For incorrect predictions, participants rated how well the AI's response matched their verbal description of the target textile on a 1–10 scale (1 = completely invalid, 10 = completely valid). It is used to identify whether incorrectness stemmed from misunderstanding human language or from limitations in the model's tactile knowledge.

For instance, if a participant described a textile as "slightly lighter, but more rough and easier to fold [in comparison to the known reference textile]," a high validity score indicated that the AI correctly interpreted these attributes (lightness, roughness, and foldability) in its predicted material. In contrast, a low score suggested semantic misalignment, such as overlooking roughness or foldability.

### 4.2. Study setup

The guessing system is hosted locally on a 13-inch MacBook Pro (Intel Core i7). The study took place in a controlled laboratory setting. Participants sat at a desk with a black box hiding the textiles from view, ensuring that judgements were based on touch alone (Fig. 2). An experimenter sat opposite, gave instructions, handed out the samples, and managed the system and interface.

At the start, participants reported their dominant hand. The target textile was always placed in that hand (e.g. "the one in your dominant hand is the target"). Participants were encouraged to manipulate both textiles freely with both hands while describing their experience. If participants became uncertain about which one was the target during exploration, they could ask the experimenter to remind them.

Textile pairs were randomly assigned by the system to ensure balanced coverage of the sample pool and to minimise order effects. Each textile was used as a target four times, with initial reference textiles rotated across the four meta-categories outlined in Section 3.3.

### 4.3. User study procedure

Each session began with a short familiarisation phase. Participants handled a single textile and described it without naming the material. They were encouraged to focus on sensory characteristics or impressions rather than material names. For example, instead of saying, "This feels like wool", they were encouraged to explain the sensations that led to this conclusion, such as "This feels like a bit furry, itchy warm, and slightly coarse".

In the main study, each participant completed two tasks. In each task they were given a reference textile (known to the AI) and a target textile (unknown to the AI, placed in the participant's dominant hand).

Participants compared the two and described the target relative to the reference. The AI system then produced a prediction.

If the prediction was correct, the task ended. If incorrect, the guessed textile became the new reference. Participants then rated the guess on two 10-point scales: validity (how well the guess matched their description) and similarity (how close it felt to the target), see Fig. 6(d) and Fig. 6(e). A new comparison round was then initiated between this updated reference and the original target.

The process continued until the AI identified the target or reached the five-attempt limit. In each task, a participant could encounter up to five rounds of comparison, involving up to six different textile pairs.

### 4.4. Participants

A total of 40 participants (30 female, 10 male; aged 18-39, mean = 25.79, SD = 4.12) were recruited for the in-person user study. None of the participants had any sensory or motor impairments that would affect their perception and handling of the textile samples. Participants had a diverse range of backgrounds, including psychology students, computer scientists, designers, artists, researchers, and university lecturers. All participants were either native English speakers or highly proficient in English. All participants provided written informed consent before participating in the study and were compensated with a gift voucher for their participation in a 30-minute study.

## 5. Results and discussion

We analysed 80 “Guess What Textile?” tasks with 362 attempts (avg 4.53 attempts per task, std = 1.41) completed by 40 participants. Below we present the overall alignment based on our three measurements and then further detail the results for textile-specific performance comparisons.

### 5.1. Overall alignment performance

#### 5.1.1. Success rates

The overall success rate is 22.5% (18/80 tasks). Among successful cases, 16 (88.9%) were completed within five attempts(avg 3.5 attempts per task). The random-guessing baseline, representing the probability of at least one correct identification within five random attempts, is 23.7%. A chi-square test showed no significant difference from chance ($\chi^2$ = 0.02, $p = .89$).

Although aggregate performance is not significantly different from chance, the distribution of successful attempts shows a non-random pattern. Successes peaked at attempt 3 (38.9% observed vs. 19.9% expected under random guessing) and clustered within attempts 2–4 (72.2%), whereas early successes (attempts 1–2) are fewer than expected (27.8% vs. 43.3%). This deviation is not statistically significant ($\chi^2$ = 4.90, $p = .297$).

#### 5.1.2. Overall validity and similarity

Validity scores (alignment with participant descriptions) averaged $M = 5.27$, $SD = 2.94$, not significantly different from the neutral midpoint of 5.5 ($p = .072, d = -0.08$). Overall, 38.7% of predictions were rated as high validity (≥ 7), 21.4% as medium (5–6), and 39.9% as low (≤ 4).

Similarity scores (perceptual closeness to the target) are slightly lower overall ($M = 4.77$, $SD = 2.78$), significantly below the neutral midpoint ($p < .001, d = -0.26$), with 31.2% rated as high, 19.7% as medium, and 49.1% as low.

Validity and similarity show a strong positive correlation ($r = 0.758$, 95% CI $[.709, 0.800]$, $p < .001$, $n = 346$). When predictions closely matched participants’ descriptions, AI was also likely to predict a guess that feels similar to the target perceptually; and vice versa, when its interpretation was poor (low validity), similarity was also typically low.

*Pattern analysis.* To interpret this relationship, predictions are grouped into four patterns based on whether each score exceeds the mid-range threshold (Fig. 7), summarised in Table 1. Most cases fall into either *High Validity, High Similarity* (26.0%), representing successful alignment where the model understood the description and produced a perceptually close output; or *Low Validity, Low Similarity* (34.4%), where the model fails to interpret the description meaningfully and perceptually.

*High Validity, Low Similarity* (4.9%) suggests that even when the model interprets participants’ words correctly and produces an output consistent with the description, perceptual similarity can still be low, likely because the description itself does not capture the qualities of the target. In contrast, *Low Validity, High Similarity* (1.2%) reflects rare cases where the model misinterprets the description yet still produces a perceptually close output. The remaining 33.5% lie near the midpoint, indicating borderline alignments.

*Correlation across task outcomes.* The validity–similarity relationship is consistent across successful and failed tasks. Correlations are statistically equivalent for *successful* ($r = 0.787$) and *failed* tasks ($r = 0.756$) as shown in Table 2. A Fisher’s $z$-test confirmed that the difference is not significant ($\Delta r = 0.031$; $z = 0.423$, $p = .672$). Mean validity and similarity scores are also statistically equivalent between the groups (all $p > .90$).

### 5.2. Textile-specific performance

To understand textile-level variation, we analysed the success rate, validity, and similarity scores for each textile. These results are depicted in Fig. 8. The results suggest that there is a significant perceptual bias across various textiles on all metrics evaluated.

#### 5.2.1. Textile-specific success rate

The success rate varies significantly across textiles, suggesting that the AI found some textiles easier to guess than others. For instance, silk satin (id 8) has achieved 100% accuracy. This could be due to various factors, on both the AI and the user sides, such as the specificity of user descriptions for certain items or inherent characteristics of the items that make them more distinguishable.

#### 5.2.2. Textile-specific validity and similarity scores

Fig. 8 displays the success rates, along with average validity and similarity scores. Average validity and similarity scores also differ considerably across textiles. We arranged the textile samples in descending order of success rate and plotted their corresponding average validity and similarity scores.

Textile-level success rates show no significant correlation with mean validity (r = 0.211, p = 0.371) or mean similarity scores (r = 0.207, p = 0.382). They exhibit considerable variance across different textiles. For instance, in Fig. 8, linen plain (ID 3, 50%) shows higher validity and similarity scores compared to silk satin (ID 8, 100%).

#### 5.2.3. Confusion matrix of AI textile predictions

We also present a confusion matrix for AI’s prediction and actual target textile in Fig. 9. The confusion matrix (Fig. 9) complements the per-textile summary in Fig. 8. Whereas Fig. 8 aggregates performance measures (success rate, validity, and similarity) at the textile level, Fig. 9 presents the same prediction data at the attempt level, showing how individual guesses are distributed across actual targets. Each cell indicates the number of times the model predicted a given textile (vertical axis) when the true target was another (horizontal axis).

The horizontal axis displays the actual textile sample IDs, while the vertical axis represents the AI’s predictions. Ideally, in an instance of perfect alignment between the AI and human judgements, the confusion matrix should exhibit its highest values along the diagonal line running from the top left to the bottom right. This perception alignment bias across textiles is further evident in the confusion matrix shown in Fig. 9, where we observe a scattering of bright spots rather than a concentrated diagonal line. The AI tends to predict linen plain (ID3), silk crepon (ID 7) and silk satin (ID 7), showing bias.

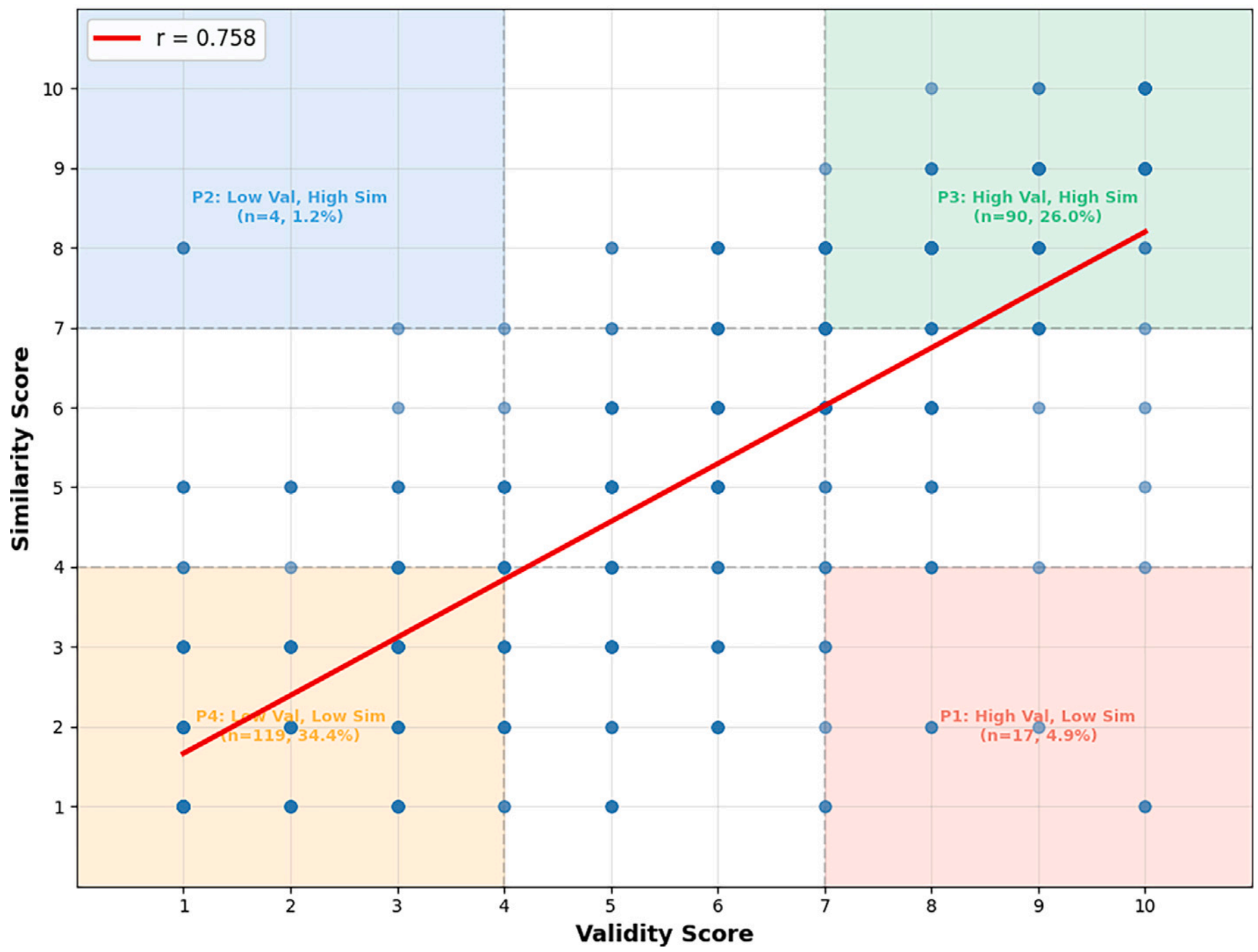

**Fig. 7.** Validity–similarity correlation with four primary outcome regions. Points represent individual attempts; coloured areas indicate distinct alignment patterns (see Table 1).

**Table 1**
Observed patterns in the relationship between validity and similarity ($N$ = 346).

| Pattern | Count (%) | Description |
|---|---|---|
| High validity, High similarity | 90 (26.0%) | Matches description; perceptually close. Successful human–AI alignment. |
| High validity, Low similarity | 17 (4.9%) | Matches description; perceptually dissimilar. AI understood participant well, but participant may not have described the target textile accurately. |
| Low validity, High similarity | 4 (1.2%) | Mismatch to description; perceptually close. Rare "lucky" matches: AI misunderstood the description but produced a perceptually similar guess. |
| Low validity, Low similarity | 119 (34.4%) | Neither matches nor close. AI's guess neither matched the participant's description nor felt similar to the target. |
| *Mid-range cases* | 116 (33.5%) | Scores near the midpoint, reflecting ambiguous or borderline alignment. |

**Table 2**
Validity–similarity relationships overall and by task outcome. Significance levels: *** $p < .001$.

| Group | n | Pearson $r$ | Validity M(SD)/Similarity M(SD) |
|---|---|---|---|
| **Overall** | 346 | 0.758*** | 5.27 (2.94)/4.77 (2.78) |
| Successful tasks | 36 | 0.787*** | 5.41 (2.69)/4.75 (2.70) |
| Failed tasks | 310 | 0.756*** | 5.25 (2.97)/4.77 (2.79) |

### 5.3. *Descriptive similarity and AI performance*

We examined whether individual differences in description influenced model performance by comparing four natural language processing (NLP) similarity measures using cosine similarity (implementation details in Appendix A):

- a *lexical* overlap measure using Term Frequency–Inverse Document Frequency (TF-IDF) vectors (Salton and Buckley, 1988);
- a *sentence-level semantic* measure using Sentence-BERT (SBERT) embeddings (Reimers and Gurevych, 2019);
- a *contextual* measure using OpenAI's `text-embedding-small` model;
- and a lexical *diversity* index (unique-to-total word ratio), representing vocabulary richness (Tweedie and Baayen, 1998).

We analysed the target textiles' descriptive similarity across the four measures. Absolute cosine values varied across methods (Table 3), as each operates in a different representational space (word frequency, sentence-level semantics, or contextual vectors) with its own internal scaling. Direct comparison of raw scores is therefore not meaningful. Rankings showed moderate agreement across approaches ($r$ = 0.45, $p$ = 0.049). The choice of similarity metric did not affect the overall results.

Correlations between descriptive similarity and AI performance were mostly weak negative and non-significant (TF–IDF $r$ = −0.29, SBERT $r$ = −0.09, OpenAI $r$ = −0.29, LexDiv $r$ = 0.09; all $p$ > 0.2). Textiles with a higher success rate are not consistently associated with higher descriptive similarity, and vice versa.

For instance, cattle leather (id 17) exhibited among the highest cosine similarities across all measures (TF–IDF = 0.223, SBERT = 0.608, OpenAI = 0.680) but achieved only average performance. Whereas silk satin (id 8) reached 100% success despite moderate similarity scores.

### 5.4. *Perceptual similarity across textiles*

We analysed pairwise similarity ratings collected during the interactive task, where participants rated the similarity of a guessed textile relative to the target on a 1–10 scale after incorrect guesses. Across the study, we obtained 137 unique textile pairs (72.1% of all possible pairings), providing sufficient coverage to reconstruct the perceptual space using multidimensional scaling (MDS) and hierarchical clustering. Full details of the similarity matrix, cluster composition, and MDS visualisations are provided in Appendix C.

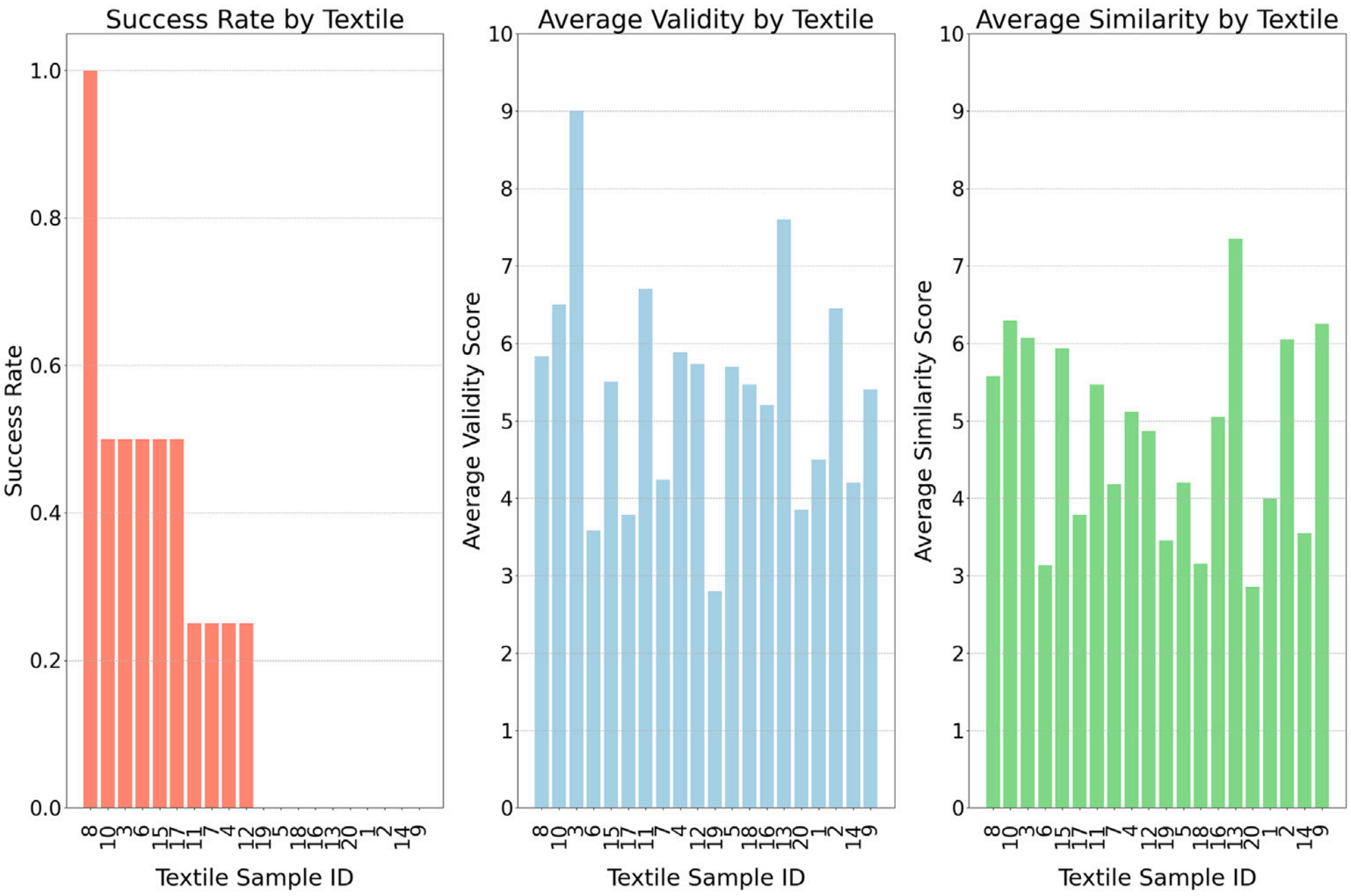

**Fig. 8.** Per-textile summary. Success rate, validity, and similarity averaged for each textile, ordered by success rate. This figure provides an aggregated view of how performance varies across textiles.

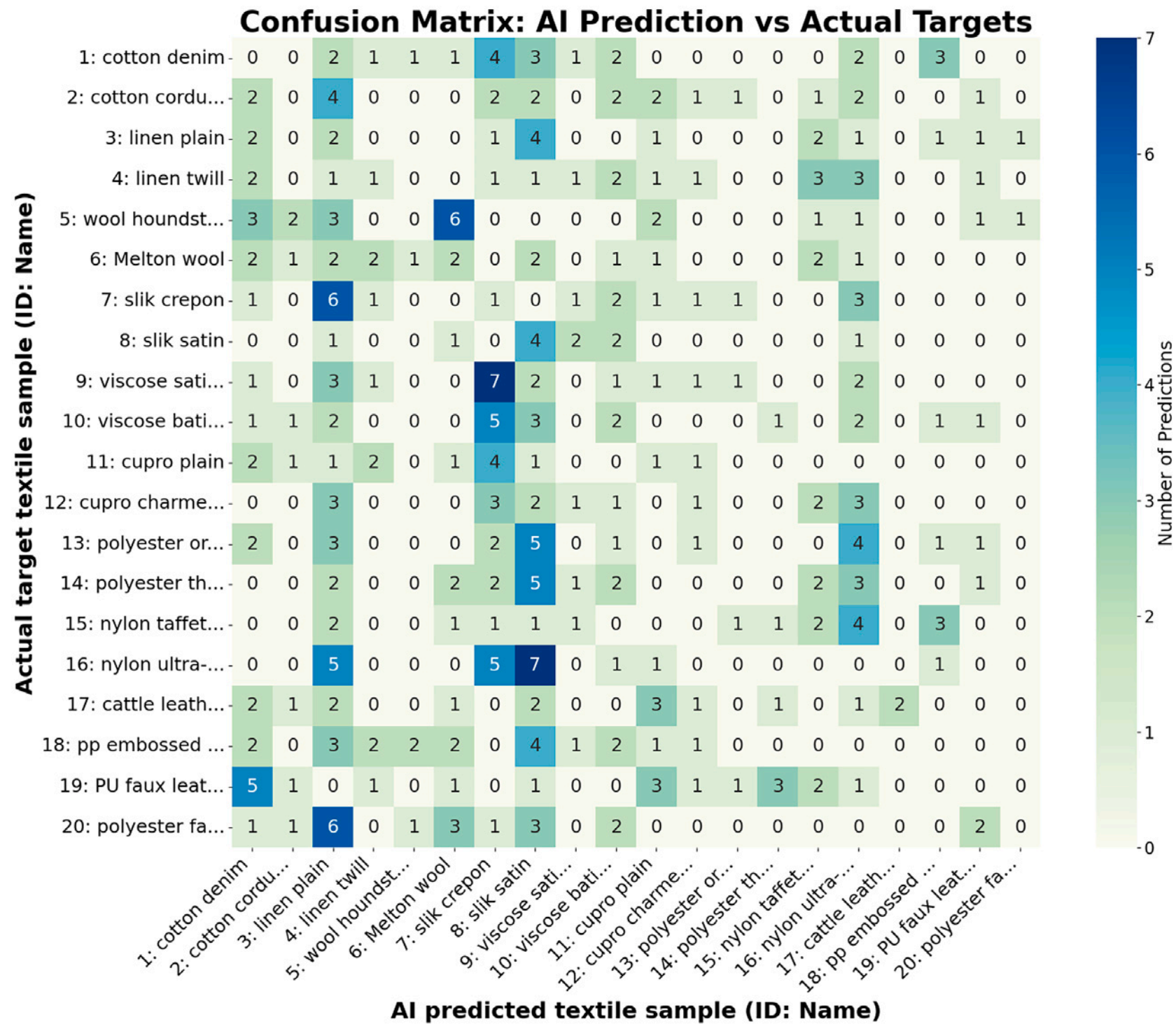

**Fig. 9.** Confusion matrix of AI predictions and actual textile targets. Each cell shows the number of prediction attempts where a specific textile was predicted (vertical axis) when another textile was the true target (horizontal axis). The diagonal represents correct identifications. Off-diagonal clusters indicate misclassifications.

Human similarity ratings spanned nearly the full scale (range: 1.0–9.6; $M = 4.43$, $SD = 2.40$, $Median = 4.00$). For example, silk_satin (ID 8) and viscose_batiste (ID 10) received the highest rating (9.60), while Melton_wool (ID 6) versus PU_faux_leather (ID 19) and silk_satin (ID 8) versus cattle_leather (ID 17) received among the lowest ratings (1.0 and 1.50, respectively). Approximately 23.4% of pairs were rated as highly similar (≥7) and 22.6% as highly dissimilar (≤2), confirming that the textile set includes both perceptually similar and clearly distinguishable materials.

Hierarchical clustering based on these ratings identified five major perceptual clusters. The five-cluster cut is determined as it balances statistical performance (peak alignment scores in Adjusted Rand Index and Normalised Mutual Information) and domain knowledge (5 meta categories used in our samples).

**Table 3**
Comparison of linguistic and semantic similarity measures by textile, ordered by AI success rate (descending). Cosine similarity was computed across three representational levels: TF–IDF captures word-level overlap, SBERT encodes sentence-level semantics, and OpenAI embeddings (`text-embedding-3-small`) represent deep contextual understanding. Higher values indicate greater internal consistency of participant descriptions, whereas lower lexical diversity values indicate more consistent wording. Bold values indicate the three highest scores per similarity column and the three lowest for lexical diversity.

| ID | Textile | TF–IDF | SBERT | OAI embedding | Lexical div | Success rate (%) |
|---|---|---|---|---|---|---|
| 8 | silk_satin | 0.099 | 0.575 | 0.614 | 0.615 | **100.0** |
| 3 | linen_plain | 0.089 | 0.476 | 0.581 | 0.492 | **50.0** |
| 6 | Melton_wool | 0.129 | 0.450 | 0.604 | 0.470 | **50.0** |
| 10 | viscose_batiste | 0.056 | 0.511 | 0.595 | 0.474 | **50.0** |
| 15 | nylon_taffeta | 0.078 | 0.479 | 0.646 | 0.591 | **50.0** |
| 17 | cattle_leather | **0.223** | **0.608** | **0.680** | 0.438 | **50.0** |
| 4 | linen_twill | 0.165 | **0.608** | **0.654** | 0.581 | 25.0 |
| 7 | silk_crepon | 0.026 | 0.433 | 0.585 | 0.579 | 25.0 |
| 11 | cupro_plain | 0.142 | 0.470 | 0.596 | 0.587 | 25.0 |
| 12 | cupro_charmeuse | 0.108 | 0.546 | 0.638 | 0.571 | 25.0 |
| 1 | cotton_denim | 0.173 | **0.636** | 0.649 | 0.487 | 0.0 |
| 2 | cotton_corduroy | 0.138 | 0.506 | **0.683** | **0.435** | 0.0 |
| 5 | wool_houndstooth | 0.128 | 0.511 | 0.595 | **0.436** | 0.0 |
| 9 | viscose_satin_stripe_organza | 0.144 | 0.583 | 0.633 | 0.641 | 0.0 |
| 13 | polyester_organza | 0.086 | 0.534 | 0.634 | 0.601 | 0.0 |
| 14 | polyester_three-layer_composite_knitting | 0.134 | 0.516 | 0.612 | 0.483 | 0.0 |
| 16 | nylon_ultra-light_and_thin_fabric | 0.113 | 0.477 | 0.646 | 0.574 | 0.0 |
| 18 | pp_non-woven_fabric | 0.136 | 0.501 | 0.661 | 0.488 | 0.0 |
| 19 | PU_faux_leather | **0.194** | 0.533 | 0.650 | 0.459 | 0.0 |
| 20 | polyester_faux_rabbit_fur | **0.169** | **0.628** | 0.634 | **0.437** | 0.0 |

#### 5.4.1. *AI semantic space and cluster alignment analysis*

To compare human and AI representations, we constructed an AI semantic space using cosine similarities between the language model's textual embeddings ($\mathcal{E}textiles = v_1, v_2, \ldots, v20$). Both human and AI similarity matrices were projected into the same low-dimensional space using MDS, and hierarchical clustering (five clusters) was applied to each space to enable direct comparison.

The AI's clustering pattern broadly mirrored human perception: 15 textiles (75%) fell within the same dominant cluster, while 5 textiles (25%) were assigned to different groups. Appendix C provides full visualisations and cluster mappings.

Overall, the results show weak-to-moderate alignment between human perceptual and AI semantic spaces (Table 4). Across all 137 rated pairs, correlations were moderate (Pearson $r = 0.441$, Spearman $\rho = 0.452$, both $p < 0.0001$), indicating that the AI captured some, but not all, human similarity trends, explaining 19.5% of the variance ($R^2 = 0.195$).

Global spatial relationships were limited: Procrustes disparity = 0.8665 (0 = perfect alignment), and pairwise distance correlations were weak (Pearson $r = 0.275$, Spearman $\rho = 0.271$). Local structure also differed, with mean nearest-neighbour overlap = 0.300 ($k = 3$), meaning on average, only 30% of each textile's 3 most similar "neighbours" were the same between human and AI spaces. In simpler terms, the textiles that people considered most alike were often different from those that the AI judged most similar. Only four textiles shared over half of their nearest neighbours. Finally, correlations between each textile's full similarity profile averaged $\rho = 0.394$, with eight textiles exceeding $\rho > 0.5$, suggesting partial but uneven alignment between human and AI judgements.

These results indicate that while the language model embeddings capture broad categorical knowledge (such as grouping wools, silks), they fail to encode the finer-grained perceptual relationships or distinctions present in human tactile discrimination. This divergence is evident in both the global spatial relationships and the local neighbourhood clustering.

### 5.5. *Variables influencing LLM alignment*

Although measured differently, Marjieh et al. (Marjieh et al., 2023) suggest that LLMs have good perceptual alignment in common modalities such as vision. For instance, they experimented on colours and observed high alignment measured by inter-rater reliability scores. We, however, observed the exact opposite for the sense of touch in textile experience. We hypothesise that this significant difference *originates from the training data*. We thus conducted a simple experiment which is to traverse the common training datasets WikiText-103 (Merity et al., 2016) and BookCorpus (Zhu, 2015). The former is a collection of articles on Wikipedia, and the latter is a large collection of novels (Zhu, 2015).

We then take a list of keywords for textiles, which are basically words and subwords from the 20 textile sample names. We also built another list of keywords that contains common colours.[3] We observed that 0.15% and 0.04% of the words in WikiText-103 and BookCorpus respectively contain colour keywords, while only 0.0033% and 0.0018% are observed for textiles. It is therefore reasonable to suggest that variations in the training data could contribute to the varying levels of alignment observed.

## 6. Discussion

### 6.1. *AI's performance in understanding touch experience descriptions*

The results show that current LLMs exhibit only a limited degree of perceptual alignment with human touch experience. The AI achieved a success rate of just 22.5% in identifying textiles from participants' descriptions, a figure statistically close to random guessing. This suggests that touch knowledge is not well encoded within text-only models. Alignment was not entirely absent: silk satin (ID 8), was always correctly identified 100% of the time, suggesting touch concepts exist within the embedding space. The internal structure may encode partial knowledge of how things feel, but this is **limited and biased**.

Validity and similarity ratings further illustrate this picture. Both averaged around neutral (M = 5.27 and 4.77), implying that participants rated the AI's interpretations as neither consistently right nor wrong, but often slightly off. In other words, the AI did not consistently capture the meaning of participants' verbal descriptions and made a valid guess. Importantly, validity and similarity correlated strongly: showing that when participants felt the AI's guess matched their words, it also tended to be perceptually closer to the target; and when one failed, both failed. This suggests that **even with non-expert descriptions, the current LLM is powerful in its linguistic understanding, and a degree of alignment may exist when relevant knowledge is encoded**.

[3] We considered "red", "orange", "yellow", "green", "blue", "purple", "pink", "brown", "black", "grey" and "white."

**Table 4**
Quantitative metrics assessing AI–human perceptual alignment.

| Analysis | Results and interpretation |
|---|---|
| *Raw similarity correlation* | Pearson $r = 0.441$, Spearman $\rho = 0.452$ ($p < 0.0001$). Moderate correspondence; AI explains 19.5% of human similarity variance. |
| *Procrustes Analysis* | Disparity = 0.8665 (0 = perfect). Indicates poor global superimposition between perceptual and semantic spaces. |
| *MDS distance matrix correlation* | Pearson $r = 0.275$, Spearman $\rho = 0.271$ ($p = 0.0001$–$0.0002$). Weak preservation of overall spatial structure. |
| *Nearest-neighbour agreement ($k = 3$)* | Mean overlap = 0.300. Only 4 textiles share more than 50% of their closest neighbours; 70% of local relations differ. |
| *Per-textile similarity profile* | Mean Spearman $\rho = 0.394$; 8/20 textiles exceed $\rho > 0.5$. Moderate alignment limited to a subset of materials. |

Importantly, participants' linguistic consistency did not drive the success. No significant correlation emerged between participants' description similarity (TF-IDF, SBERT, OpenAI embeddings) and the AI's success rate. Several materials that participants described as highly similar did not lead to a higher success rate, and in some cases, even worse outcomes (e.g. AI failed in all faux fur tasks). This highlights two key points: first, when participants used similar descriptions, it showed that they largely agreed on how those materials feel and how they can be described; and second, the AI did not share this agreement. In other words, success depended not on consistent wording, but on whether the LLM had learned to represent and understand the relative differences in how materials feel. That is, whether its embeddings capture concepts of touch.

LLM's performance also varied significantly across textiles. Materials such as *silk satin* (ID 8), and *cattle leather* (ID 17), achieved higher success, validity, and similarity scores, whereas others like cotton denim, faux leather, and faux fur failed entirely. This was unexpected: we hypothesised that faux fur and faux leather, having distinctive textures and being consistently described by participants, would be easier for the AI to recognise. Instead, the model failed to capture how they feel differently. Concepts such as leather being thick, resistant to folding or stretching, and having two distinct surfaces may not yet be well represented in the embedding space.

The model's internal structure helps explain why its understanding of touch remains limited (Section 5.4). LLM tended to cluster textiles that are categorically related, such as cotton and linen (both natural fibres), rather than by how similar they actually feel to humans. We found only four textiles shared more than half of their closest neighbours between the human and AI spaces, showing a clear mismatch. Ideally, if perceptual alignment existed, the LLM's embedding space could mirror the geometry of human perceptual space, grouping materials not only by technical classification but also by how they feel. Just as cats and lions cluster because they look alike, a model should place silk satin and viscose batiste together because they feel alike. Current LLMs exhibit some sensory alignment in other modalities (Marjieh et al., 2023), but this pattern reflects a representational bias: they may lack a touch dimension grounded in physical, embodied experience—at least for textiles.

In short, current LLMs do not yet encode enough about how things feel. Their internal representations cluster materials by categorical or linguistic similarity rather than by perceptual similarity, showing that a "touch space" has not yet emerged. Nonetheless, the few clear successes (e.g. silk satin at 100%) indicate that perceptual alignment could be achievable with further model development. Such alignment appears to emerge only where language has already done the heavy lifting, when strong verbal associations exist in textual corpora, as with silk satin. The model may have learned how to describe individual materials, but it has not learned how these sensations relate across different dimensions of touch. For instance, it may recognise that concrete is hard, silk is soft, and leather is hard. Yet if asked "which is harder, leather or concrete?", the model would likely choose concrete purely based on linguistic frequency rather than an understanding of tactile hardness. Humans, on the other hand, can interpret that leather may be described as both hard and soft depending on which aspect of touch is meant—hard in terms of firmness, but soft and smooth in terms of surface texture. The model, however, lacks this multidimensional understanding and treats each material as an isolated description rather than as part of a connected perceptual space.

### *6.2. Why touch remains hard for large language models*

The challenge of touch alignment arises from the modality itself. Unlike vision or sound, which have established descriptive vocabularies and quantitative metrics(e.g. RGB values, frequency scales), touch is inherently multidimensional (e.g. texture, temperature, and pressure), and is highly context-dependent. There is no universal sensor or scale for "softness" or "smoothness", and many tactile descriptors are metaphorical, such as "buttery" or "crisp". For example, the word "soft" can describe very different sensations: a smooth surface, a flexible structure, or a gentle temperature. These distinctions are obvious to humans as common sense but are rarely captured in text (Richardson and Heck, 2023). Prior HCI research has shown that human tactile descriptions vary and are often ambiguous (Obrist et al., 2013; MacLean et al., 2017). Because of this complexity, touch is rarely verbalised in consistent or systematic ways, limiting how well it can be captured in language and, in turn, learned by language models.

**Touch is rarely written, which limits how well language models can learn it**. Everyday text seldom describes how things feel; tactile qualities are treated as shared commonsense rather than explicitly narrated (Klatzky et al., 1985; Gabriel et al., 2004). A novel may mention someone wearing denim jeans without describing the fabric's coarse weave or stiffness, yet silk or velvet are more likely to be portrayed as soft, smooth, or delicate because of their cultural and emotional associations. Product reviews similarly praise silk's cool drape or flowing feel but rarely comment on the physical texture of cotton denim or polyester faux fur. This in-context pattern means that tactile knowledge may enter language only when it carries symbolic or affective meaning (Feldman and Narayanan, 2004). Our corpus analysis supports this hypothesis: in major pre-training datasets (e.g. WikiText-103, BookCorpus), tactile and textile terms appear much less often than colour words. Such evidence confirms that touch remains underrepresented in text-only corpora, leaving large gaps in the linguistic input that LLMs rely on.

Another important concern is who provides the language that bridges the gap between human living experiences and AI understanding. Our study intentionally recruited non-expert participants to evaluate LLMs using the everyday descriptions typical of real users. This reflects how current AI systems powered by LLMs operate in practice, where conversational assistants, recommendation agents, and creative tools must interpret ambiguous, unstructured, and context-rich non-expert language. Expert terminology (e.g. "warp-weft balance") may differ greatly from everyday expressions ("itchy but cosy", "soft yet thin"). This variation is not incidental but central to perceptual alignment: an LLM should mediate between professional and everyday semantics, not privilege one over the other. Models that only handle expert language can remain both socially and functionally limited. Examining how LLMs respond to everyday language therefore provides a more practical and forward-looking measure of perceptual alignment, one that better reflects real-world interaction and highlights the inclusive challenges of human–AI communication.

### 6.3. Implications for human–AI perceptual alignment

Although this study uses textiles as case study, its implications extend well beyond textile touch. Yet, even this limited alignment carries important implications for how we conceptualise and design future human–AI systems. As AI systems increasingly mediate interactions with the physical world, from recommending clothing to controlling robotic manipulators, **perceptual misalignment poses a serious barrier to usability and trust**. If an AI cannot interpret how something feels, it cannot meaningfully assist in contexts where touch is central to decision-making or experience.

First, alignment cannot be assumed to generalise across modalities. Generative models, like DALL-E for image (Ramesh et al., 2021), MusicLM for sound (Agostinelli et al., 2023), and Sora for video (Brooks et al., 2024), demonstrate recent advancements in visual and audio alignment. In contrast, touch is difficult to digitize (Zhong and Obrist, 2025) and difficult to describe consistently (MacLean et al., 2017; Obrist et al., 2013). Our results reinforce this distinction: LLMs have limited tactile knowledge because the modality is underrepresented. Perceptual alignment must therefore be treated as modality-specific; each sense demands distinct forms of grounding, representation, and evaluation. A system aligned for visual perception may still be perceptually “blind” to touch, as exemplified by recent work in cross-modal object detection (Gao et al., 2023), cross-modal texture prediction (Hassan, 2023), and simulation-guided perception modelling (Lim et al., 2025).

Second, our findings point to the importance of linguistic diversity in alignment evaluation. Everyday tactile language is fluid and metaphorical, differing sharply from expert or technical discourse (Obrist et al., 2013; MacLean et al., 2017). For alignment to be meaningful in real-world use, where people rarely speak in controlled, domain-specific terms, models must be able to interpret and respond to ambiguous, everyday phrasing. Most existing work relies on synthetic labels generated by LLMs (Fu et al., 2021) for scalability; however, this risks a “garbage in, garbage out” problem (Shumailov et al., 2024). We therefore **call for participatory large-scale datasets** and evaluation protocols that reflect how people actually talk about touch, not just how experts or textbooks describe it.

Third, ethical safety constraints currently further limit the development of touch in language models. Fu et al. (2021) reported that attempts to fine-tune LLaMA2 (Touvron et al., 2023) for touch understanding were blocked by safety filters: the model refused to answer questions about how things feel. We encountered similar restrictions when using GPT-based models (e.g. GPT-4) to assess descriptive similarity in Section 5.3. We were unable to complete the full experiment, as the model declined to respond to 12.5% of the tasks due to safety constraints. These mechanisms are designed for ethical reasons, such as preventing misuse of generative models given that “touch” often appears in sexual contexts online (Callister et al., 2012). However, such blanket restrictions inadvertently hinder research on legitimate touch research and the development of safe, embodied understanding. Future systems therefore require finer-grained safeguards that can distinguish between harmful and neutral uses of tactile description, enabling progress without compromising safety. More open discussion is also needed on how models handle and moderate touch-related language, including how boundaries are defined, who sets them, and how they shape what AI can learn about embodied experience.

Last, these implications extend beyond tactile perception. Touch shows where AI still struggles to understand how people experience the physical world. Addressing this gap will enrich not only AI’s perceptual competence but also its social and ethical integration into everyday life. As AI systems increasingly mediate material and sensory experiences, from fashion and design to healthcare and assistive technologies, achieving robust perceptual alignment becomes essential to ensuring that human–AI interaction remains intuitive, inclusive, and trustworthy. Understanding how models represent, constrain, and ethically handle embodied concepts such as touch will be key to building AI systems that can engage with the physical world in genuinely human-centred ways.

### 6.4. Limitations and future work

This study has several limitations that should be acknowledged. First, our experiments focused on textiles as a representative domain for touch. While this provides a controlled and interpretable testbed, it does not capture the full diversity of touch experiences found in other materials or everyday interactions. Future research could extend this approach to a wider range of objects, surfaces, and contexts, including temperature, weight, and compliance, to better understand how language models handle different tactile dimensions (Lim et al., 2025; Hassan, 2023).

Second, the sample size and linguistic diversity of participant descriptions were limited. Our data were collected in English and primarily reflected non-expert, everyday language. Future studies could examine multilingual and cross-cultural perspectives on tactile descriptions, as meanings of words such as *soft*, *warm*, or *rough* vary significantly across languages and cultures.

Despite these limitations, this work provides a first step towards understanding how language models represent touch and where their perceptual limits lie.

Future research can investigate how multimodal large language models (MLLMs) might enhance human–AI perceptual alignment, extending beyond text-based understanding to richer, embodied representations of touch. Integrating sensory data such as visual, auditory, or haptic information (Fu et al., 2021; Peng et al., 2023; Lim et al., 2025) could allow models to reason more effectively about how materials feel and interact with the physical world with richer, more embodied representations. As AI systems become increasingly embedded in everyday devices and services, exploring multimodal human–AI interaction will be essential for intuitive and perceptually aligned design.

At the same time, these advances depend on the quality of the underlying language models. Because current LLMs form the foundation for most multimodal systems, meaning that their existing linguistic biases and structural limitations can easily propagate upward. Without stronger alignment at the foundational level, multimodal extensions risk inheriting existing biases, a “garbage in, garbage out” (Shumailov et al., 2024) problem that limits perceptual learning.

There also remains a fundamental lack of large-scale data on touch. Tactile experiences are rarely written about or discussed in everyday text, leaving major gaps in the corpora used to train language models. Future work should therefore focus on collecting and curating large-scale, diverse datasets that capture how people describe and interpret tactile sensations in natural contexts, such as Seifi’s work on vibration libraries (Seifit et al., 2015). Such efforts would provide a foundation for models to learn the linguistic and perceptual dimensions of touch more effectively.

Last, safety restrictions on generative models limited both experimentation and evaluation. Some models refused to generate or rate tactile descriptions due to content filters, preventing a complete analysis. They highlight an important area for further discussion. Future research should explore how ethical safeguards can be maintained without constraining legitimate research on touch, embodiment, and perception. More open dialogue between AI developers, HCI researchers, and ethicists will be crucial to developing responsible approaches to modelling sensory experience.

By combining multimodal learning, inclusive datasets, and ethically aware design, future research can move closer to developing AI systems that understand how things feel.

## 7. Conclusion

In this paper, we explored human–AI, specifically human-LLM, perceptual alignment using the textile hand concept. We developed a “Guess What Textile” interactive task and conducted an in-person user study with 40 participants. Our results suggest some level of perceptual alignment; however, we observed a bias in the LLM across various textiles. We observed that there is significantly greater alignment with human perception for certain textiles compared to others (e.g. silk satin versus cotton denim).

Overall, current LLMs show only a limited and biased understanding of touch. The models’ internal representations cluster by linguistic or categorical similarity rather than by perceptual proximity, indicating that the “touch space” has not yet emerged.

In this exploratory work, we highlight that LLMs are still in their infancy concerning sensory judgment, particularly in the realm of tactile perception. Before they are widely used in embodied AI systems such as robots, assistive devices, or design tools, their perceptual alignment must be improved. Without better foundational representations, multimodal extensions risk repeating the same biases—the “garbage in, garbage out” problem that undermines reliable performance in the physical world. Future models that integrate multimodal sensory data, including visual, auditory, and haptic inputs, could explore relational representations of how things feel.

These findings highlight both the promise and the challenge of achieving human–AI perceptual alignment. Touch represented in language alone reaches its limits: it is rarely verbalised, often metaphorical, and deeply context-dependent. Addressing this gap will require multimodal approaches that integrate physical grounding and sensory experience with language. It also calls for more open discussion about how ethical safeguards shape what models can learn about touch and embodied experience.

## CRediT authorship contribution statement

**Shu Zhong:** Writing–original draft, Visualization, Software, Methodology, Investigation, Formal analysis, Data curation, Conceptualization. **Elia Gatti:** Writing–review & editing, Supervision, Methodology, Conceptualization. **Youngjun Cho:** Writing–review & editing, Supervision, Funding acquisition. **Marianna Obrist:** Writing–review & editing, Supervision, Methodology, Funding acquisition, Conceptualization.

## Declaration of competing intrest

The authors declare the following financial interests/personal relationships that may be considered potential competing interests:

Shu Zhong reports that financial support was provided by UK Research and Innovation. If there are other authors, they declare that they have no known competing financial interests or personal relationships that could have appeared to influence the work reported in this paper.

## Acknowledgments

This work was funded by the Engineering and Physical Sciences Research Council (EP/V011766/1) National Interdisciplinary Circular Economy Research (NICER) programme for the UKRI Interdisciplinary Circular Economy Centre for Textiles: Circular Bioeconomy for Textile Materials.

## Appendix A. Textile description template

In our study, the descriptions are generated by humans and SOTA LLMs.

In handcrafted textile description, the template is:

[Textile Name] is [characteristic]. [Additional information from sample book]. [Composition], is a [fibre category type] produced by [raw material], [characteristic of fibre]. [Fabric] is [produce method] and [characteristic]. [Textile name] is commonly used for [application].

The textiles are from 4 meta categories with 4 special cases, we provide their descriptions used in our study.

***Natural fibres***:

Pure cotton denimPure cotton denim is a heavy, non-elastic, coarse, thick, and textured textile. Cotton, as a natural plant fibre, is known for its durability and breathability. It has the ability to absorb and release moisture rapidly, although it is prone to shrinking and wrinkling. Denim is a woven twill known for its strength and durability, which results in a coarse texture. Denim’s stiffness allows garments to hold their shape well, but it softens with washing and wear. Cotton denim is a popular and casual textile commonly used to make skirts, jackets, and jeans.

Fine cotton corduroyFine cotton corduroy is made from 100% cotton with a cut pile weave that gives it a unique texture. Cotton, as a natural plant fibre, is known for its durability and breathability. It has the ability to absorb and release moisture rapidly, although it is prone to shrinking and wrinkling. Corduroy is thick, durable, and cosy due to its soft, corded, cut-pile brushed nap on the surface and twill weave at the back. Cotton corduroy is a common choice for jackets, trousers, work shirts and home decor.

Plain linenPlain linen has a crisp, lightweight texture with a natural linen feel, making it ideal for staying cool in warm weather. Linen, obtained from flax or linseed plants, is stiff, crisp, cool and breathable with little elasticity. Plain weave fabric has a simple interlacing warp thread, producing a firm fabric. Plain linen is commonly used for summer clothing, baby clothing and bedding.

Linen 2 × 2 twillLinen 2 × 2 twill is a durable, crisp, and sturdy textile with warp-faced steps on the surface. Linen, obtained from flax or linseed plants, is stiff, crisp, cool and breathable with little elasticity. The two-by-two twill weave is sturdy and firm and has a slight water resistance due to its tight weave. Twill linen is commonly used for outerwear, trousers and aprons.

***Animal fibres***:

Wool houndstoothWool houndstooth, a classic woven tweed textile, is known for its distinctive, broken check pattern, resembling a dog’s tooth. Wool is a natural animal fibre from sheep that is crimped and wavy. Before it is treated, it has a lofty and slightly greasy hand feel, which may irritate the skin. Wool can retain air and warmth. Houndstooth is a rough fabric with an intricate weaving pattern that features broken checks and abstract, geometric shapes that help retain heat and keep warmth. Wool houndstooth is commonly used for jackets, coats, gloves and other tailored garments.

Melton woolMelton wool is a remarkably warm and comfortable textile with a thick, quasi-felted texture. Wool is a natural animal fibre from sheep that is crimped and wavy. Before it is treated, it has a lofty and slightly greasy hand feel, which may irritate the skin. Wool can retain air and warmth. Melton woollen fabric is made by weaving wool into twill, then felting and heavily brushing it to give it a smooth, almost non-fuzzy finish. It has excellent insulating properties. Melton wool is a popular choice for coats, jackets, blankets, and cold-weather accessories.

Silk creponSilk crepon, also known as crinkled silk, is a very lightweight, semi-sheer, pleat-like textile with a vertically rippled texture. Silk is a natural animal fibre derived primarily from moths and it is the only natural filament fibre that is soft, lustrous, and smooth to the touch. Crepon has a loose, coarse weave and a pleat-like texture that can be pressed flat. Silk crepon is commonly used for evening wear, blouses, dresses and scarves.

**Table A.5**
List of the textiles sample used in the study.

| label_int | name | fibre | fabric | label_int | name | fibre | fabric |
|---|---|---|---|---|---|---|---|
| 1 | cotton_denim | cotton | denim | 11 | cupro_plain | cupro | plain |
| 2 | cotton_corduroy | cotton | corduroy | 12 | cupro_charmeuse | cupro | charmeuse |
| 3 | linen_plain | linen | plain | 13 | polyester_organza | polyester | Organza |
| 4 | linen_twill | linen | twill | 14 | polyester_three-layer composite knitting | polyester | three-layer_ composite_knitting |
| 5 | wool_houndstooth | wool | Houndstooth | 15 | nylon_taffeta | nylon | taffeta |
| 6 | Melton_wool | wool | Melton | 16 | nylon_ultra-light and thin_fabric | nylon | ultra-light |
| 7 | Silk_crepon | silk | crepon | 17 | cattle_leather | leather | leather |
| 8 | slik_santin | silk | satin | 18 | pp_non-woven-fabric | pp | non-woven |
| 9 | viscose_satin_stripe_organza | viscose | satin_stripe_organza | 19 | PU_faux_leather | PU | fuax_leather |
| 10 | viscose_batiste | viscose | batiste | 20 | polyester_faux_rabbit_fur | polyester | fuax_rabbit_fur |

Silk satinSilk satin is a luxuriously smooth, soft, glossy, and non-elastic textile. Silk is a natural animal fibre derived primarily from moths and is the only natural filament fibre that is soft, lustrous, and smooth to the touch. The satin fabric is smooth and lustrous with a silky front, while the back is dull. Its slippery texture makes it difficult to handle. Silk satin is commonly used for gowns, dresses, blouses, linings, scarves and bedding.

***Regenerated fibres*:**

Viscose satin stripe organzaViscose satin stripe organza blends the structured transparency of organza with the glossy sheen of satin to create a light and airy fabric. Viscose is a regenerated fibre made from wood pulp with a silky texture, often called artificial silk. Viscose can be less durable than other regenerated rayon. Satin stripe organza with added satin stripes on the organza provides subtle texture and visual interest. Viscose satin stripe organza suits elegant dresses, skirts, and decorative home textiles.

Viscose batisteViscose batiste is a lightweight, soft, breathable, smooth textile that wrinkles easily. Viscose is a regenerated fibre made from wood pulp with a silky texture, often called artificial silk. Viscose can be less durable than other regenerated rayon. Batiste is a lightweight, durable, semi-sheer plain weave fabric with a fine, smooth, and soft texture. Viscose batiste is commonly used for shirts, skirts and dresses.

Plain cuproPlain cupro is a silky drape textile with a smooth texture and a high thread count. Cupro is a regenerated fibre obtained from cotton linter with a smooth, silky texture and a shiny surface. It is also breathable, hypoallergenic, anti-static, stretchable, and capable of regulating body temperature, making this fibre an eco-friendly alternative to silk. Plain weave fabric has a simple interlacing warp thread resulting in a firm fabric. Plain cupro is commonly used for dresses, skirts and tops.

Cupro charmeuseCupro charmeuse is a lightweight, thin, satin-like textile that is lustrous in the front and dull at the back. Cupro is a regenerated fibre obtained from cotton linter with a smooth, silky texture and a shiny surface. It is also breathable, hypoallergenic, anti-static, stretchable, and capable of regulating body temperature, making this fibre an eco-friendly alternative to silk. Satin crepe, a reversible fabric, has a slightly silky front and a crepe-textured back. Cupro charmeuse combines the breathability and comfort of cupro with the glossy sheen of satin, making it perfect for high-end fashion pieces.

***Synthetic fibres*:**

Polyester organzaPolyester organza is a lightweight, sheer textile with a stiff drape and crisp texture. Polyester is a common synthetic fibre that is soft and incredibly durable, making it a popular choice for clothing. It drapes easily, holds the shape of an item of clothing well, and can hold pleats created by heat. Organza fabric is known for its supportive nature, sheer quality, and lightweight feel, with a stiff and crisp texture. Polyester organza is often used in dresses, bridal wear, evening gowns and packaging.

Polyester Triple-layer composite knitThe polyester triple-layer composite knit is a functional textile that is thick, windproof and breathable. Polyester is a common synthetic fibre that is soft and incredibly durable, making it a popular choice for clothing. It drapes easily, holds the shape of an item of clothing well, and can hold pleats created by heat. The triple-layer knit has distinct functions for each layer; the outer layer may be water- and wind-resistant, and the inner layer offers insulation and moisture-wicking qualities. When combined, the layers provide a balance between wind protection and breathability. Polyester triple-layer composite knit is commonly used in outerwear and sportswear.

Nylon taffetaNylon taffeta is a sturdy, durable textile that rustles when moved. Nylon, a synthetic fibre, has a smooth and soft texture, but it is also highly durable, water-repellent, wrinkle-resistant, and easily attracts static electricity. Taffeta is a stiff, crisp, smooth fabric that will make a rustling sound when rubbed. Nylon taffeta is commonly used for gowns, jackets, sportswear, sleeping bags and handbags.

Nylon ultralight thin fabricUltralight nylon thin fabric is feather-light, smooth and airtight. Nylon, a synthetic fibre, has a smooth and soft texture, but it is also highly durable, water-repellent, wrinkle-resistant, and easily attracts static electricity. Ultralight nylon fabric weighs just 51g/m2 and is windproof and water resistant with a plain weave. Nylon fabric is commonly used for skin-friendly trench coats and lightweight down jackets.

***Specialty fabrics*:**

Cattle leatherCattle leather is a durable, thick and flexible material made from cattle hide with a unique animal skin texture on the front and a smooth, napped finish on the back. Cattle leather is a special case in animal fibre made from tanned cattle hide; leather fibre gives it resistance to tearing and abrasion. Cattle leather is widely used in jackets, shoes and accessories.

Polypropylene embossed non-wovenPolypropylene embossed nonwoven fabric is lightweight and has good air permeability and filtration properties; it is made of polypropylene fibres bonded together without weaving. Polypropylene (PP), also known as the ’steel of plastics’, is a synthetic fibre that comes in long and short fibres: long for silk-like and knit fabrics, short for carpets and non-wovens. It’s lightweight, with low moisture absorption and heat resistance, and offers high strength, elasticity, and durability. Embossed non-woven fabric is neither woven nor knitted but formed by chemical or mechanical treatment that is water-absorbent and waterproof. PP non-woven is commonly used in home decor and storage packaging.

PU faux leatherPU faux leather is a synthetic textile with a PU coating over a polyester warp-knitted base that mimics the look and feel of genuine leather while having a lychee-patterned texture. PU is a synthetic fibre, also known as polyurethane, and it is a synthetic material used to create a leather-like fabric. PU faux leather is a popular choice among consumers who prioritise vegan and eco-friendly products for clothing, bags, and home decor.

Polyester faux rabbit furSynthetic polyester faux rabbit fur is created to mimic the softness and texture of real rabbit fur. It offers furry,

**Table B.6**
Average description length and total word volume per textile, ordered by success rate (descending). "Avg Len ± SD" represents the mean and standard deviation of description length (in words). Bold values indicate the three highest per column.

| ID | Textile | Avg Len ± SD | Total Words | Success Rate (%) |
|---|---|---|---|---|
| 8 | silk_satin | 44.9 ± 17.0 | 411 | **100.0** |
| 10 | viscose_batiste | 33.1 ± 12.4 | 538 | **50.0** |
| 3 | linen_plain | 33.4 ± 19.3 | 445 | **50.0** |
| 6 | Melton_wool | 40.8 ± 9.4 | 562 | **50.0** |
| 15 | nylon_taffeta | 35.5 ± 20.4 | 477 | **50.0** |
| 17 | cattle_leather | 65.1 ± 26.4 | 848 | **50.0** |
| 4 | linen_twill | 49.0 ± 28.9 | 728 | 25.0 |
| 7 | silk_crepon | 40.8 ± 17.1 | 580 | 25.0 |
| 11 | cupro_plain | 39.4 ± 31.2 | 452 | 25.0 |
| 12 | cupro_charmeuse | 38.6 ± 14.7 | 509 | 25.0 |
| 1 | cotton_denim | 44.9 ± 40.9 | 714 | 0.0 |
| 2 | cotton_corduroy | 38.6 ± 19.8 | 616 | 0.0 |
| 5 | wool_houndstooth | 26.6 ± 8.6 | 438 | 0.0 |
| 9 | viscose_satin_stripe_organza | 26.8 ± 10.5 | 454 | 0.0 |
| 13 | polyester_organza | 24.6 ± 14.9 | 413 | 0.0 |
| 14 | polyester_three-layer_composite_knitting | 38.3 ± 23.9 | 640 | 0.0 |
| 16 | nylon_ultra-light_and_thin_fabric | 28.1 ± 16.5 | 462 | 0.0 |
| 18 | pp_embossed_non-woven_fabric | 36.2 ± 17.5 | 599 | 0.0 |
| 19 | PU_faux_leather | 56.8 ± 24.0 | 949 | 0.0 |
| 20 | polyester_faux_rabbit_fur | 63.1 ± 25.1 | 1038 | 0.0 |

warmth and luxury. Polyester faux rabbit fur is commonly used for coats, jackets, and home décor.

## Appendix B. Participants' description similarity analysis

All textual analyses were implemented in Python 3.11 using standard natural language processing libraries.

**Lexical similarity (TF-IDF)**

Participant descriptions were tokenised and vectorised using scikit-learn's `TfidfVectorizer`. Stop words were removed using a shared list across all analyses, based on NLTK's English stop-word set extended with frequent task terms (*feel*, *feels*, *like*, *one*, *two*, *reference*, *target*, *textile*, *fabric*, *material*, *sample*). For all three similarity measures (TF–IDF, SBERT, and OpenAI), cosine similarity was computed between every pair of participant descriptions within each textile (i.e. all pairwise comparisons, $n(n-1)/2$ per textile), and the resulting values were averaged to obtain a single internal similarity score per textile.

**Semantic similarity (Sentence-BERT)** Sentence-level embeddings were computed using the pre-trained `all-MiniLM-L6-v2` model from the *sentence transformers* library (384 dimensions). Mean cosine similarity was calculated across all description pairs for each textile.

**Contextual similarity (OpenAI embeddings)** Contextual representations were obtained using OpenAI's `text-embedding-small` model (1,536-dimensional embeddings). Average within-text cosine similarity served as the contextual similarity measure. **Lexical diversity** Vocabulary richness was defined as the ratio of unique to total *content* words after applying the same stop-word list; tokens were lowercased prior to counting to normalise case variants. This index reflects diversity (higher = more varied wording), not similarity.

**Verbal description characteristics**

We also computed complementary linguistic metrics to characterise participants' verbal behaviour during the task. We calculated description the length (average number of words per description) and its variability in Table B.6. Across all 362 descriptions, the mean length is 40.7 words (SD = 21.5), indicating variation in how much participants spoke (coefficient of variation = 0.53). Within-session variability averaged 9.4 words (range = 0–29.3), showing that participants adjusted the amount of verbal detail across attempts.

Comparing successful (n = 18) and failed sessions (n = 62), successful interactions involved slightly longer descriptions (46.5 ± 19.9 words) than failed ones (39.1 ± 21.9 words), though the difference was not significant ($t = 1.30$, $p = 0.196$). At the textual level, higher total word counts did not predict greater AI accuracy ($r = 0.08$, $p = 0.15$), indicating that the amount of verbal information alone did not explain performance.

**Correlation analysis** For each textile, mean similarity and diversity scores were correlated with AI success rates using Pearson's $r$ (two-tailed, $\alpha = 0.05$) to test whether descriptive consistency predicts model accuracy.

## Appendix C. Perceptual similarity across textiles

### C.1. Pairwise perceptual similarity ratings

During the interactive guessing task, participants rated the similarity of a guessed textile relative to the target on a 1–10 scale after incorrect guesses. This resulted in ratings for 137 unique pairs (72.1% coverage out of all 190 possible). Ratings were averaged across participants to construct a $20 \times 20$ perceptual similarity matrix (Fig. C.10). Warm-coloured blocks indicate perceptually similar textiles, while cooler colours highlight strongly dissimilar ones. White cells denote pairs without ratings.

Human similarity ratings spanned the nearly full scale (1.0–9.6; $M = 4.43$, SD = 2.40, Median = 4.00). Around 23.4% of pairs were rated highly similar (≥7), while 22.6% were rated very low (≤2), confirming that the textile set covers both perceptually similar and clearly distinguishable materials. Hierarchical agglomerative clustering with Ward linkage revealed five coherent perceptual groups. Within-cluster similarity ($M = 6.30$) was substantially higher than between-cluster similarity (3.74), producing a separation ratio of 1.68 and a large effect size (Cohen's $d = 1.21$).

### C.2. Perceptual vs. AI semantic space

AI-derived semantic similarity was computed from language model embeddings ($\mathcal{E}_{textiles} = \{v_1, v_2, \ldots, v_{20}\}$) using cosine similarity. Metric multidimensional scaling (MDS, two dimensions) projected both human and AI similarity matrices into comparable low-dimensional spaces. The human space (Fig. C.12) revealed five perceptual clusters, while the AI space (Fig. C.13) showed a more compressed structure with partially overlapping groups. Some high-level structures (e.g. silk and viscose grouping) were shared, but many textiles were repositioned, mirroring our finding that the language model may not fully capture tactile information yet.

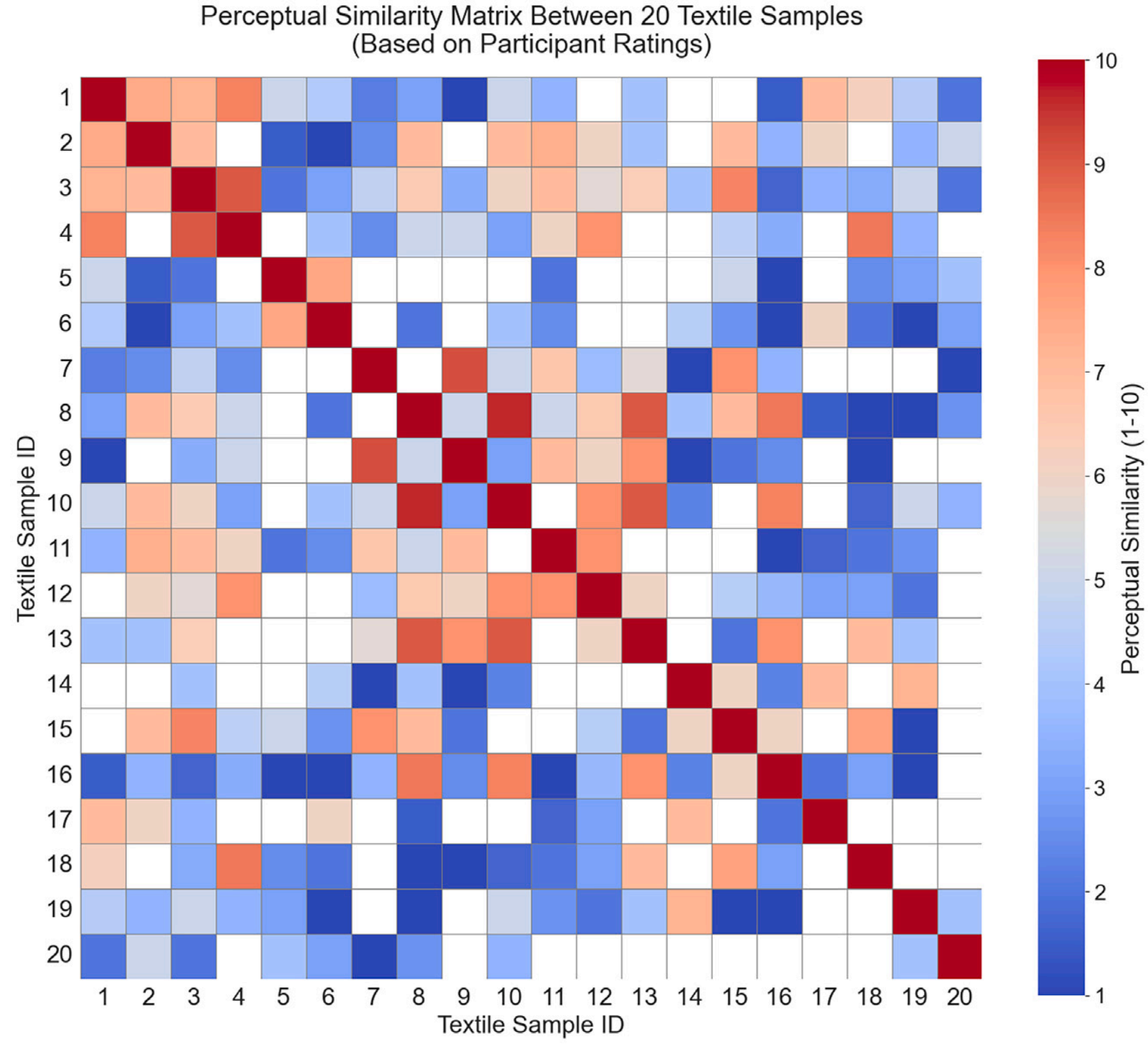

**Fig. C.10.** Perceptual similarity matrix of 20 textiles (1–10 scale). Warm colours indicate higher perceived similarity. Distinct red blocks reveal coherent perceptual families, forming the basis for subsequent MDS and clustering analyses.

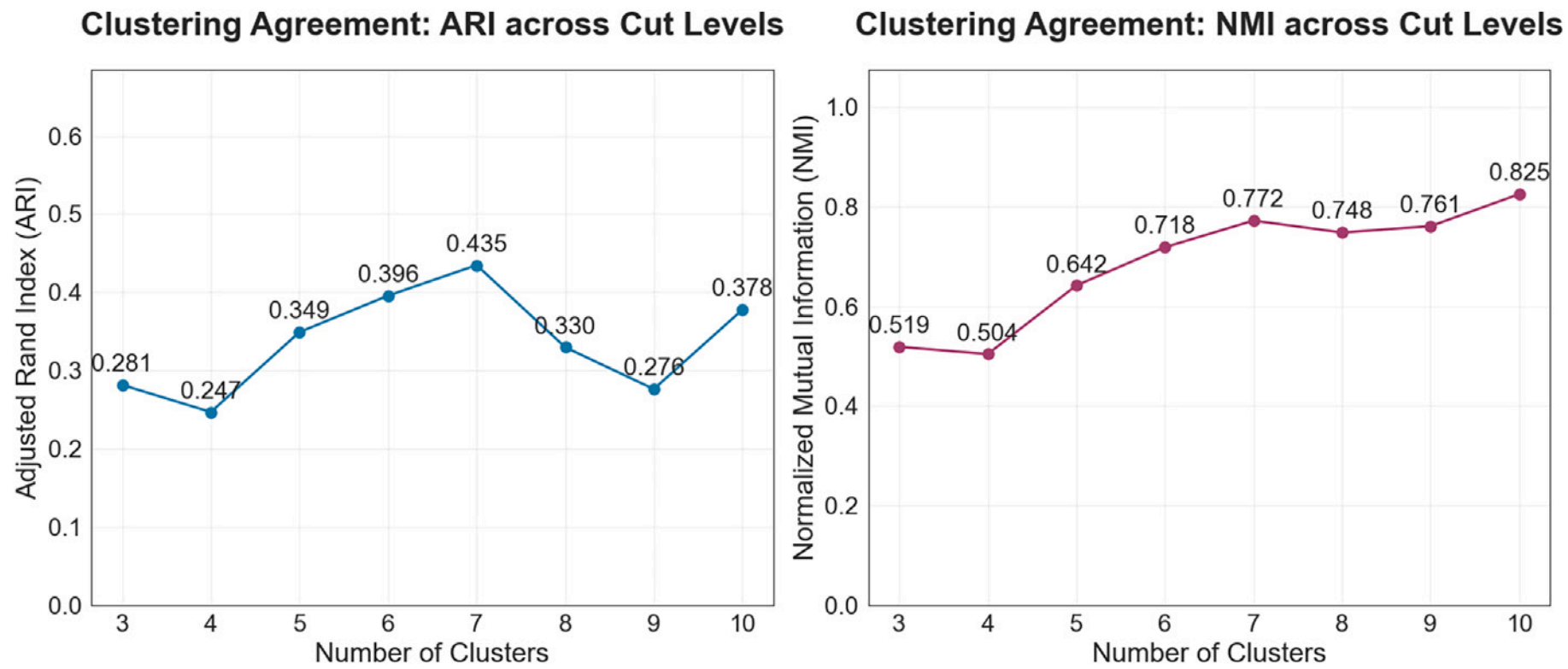

**Fig. C.11.** Clustering agreement across cut levels. Left: Adjusted Rand Index (ARI) showing alignment strength relative to random baseline. Right: Normalised Mutual Information (NMI) relative to an independent baseline. The $k = 5$ solution offers a strong balance between alignment performance and interpretability.

We selected $k = 5$ clusters to balance statistical performance and domain relevance. Adjusted Rand Index (ARI) [4] and Normalised Mutual Information (NMI) showed optimal alignment between human and AI groupings at this level (Fig. C.11), which also mapped naturally onto four major fibre categories (natural, animal, regenerated, synthetic) plus a fifth special category for non-fibre materials.

Overall, the analyses show that human perceptual similarity spans a rich continuum of textile relationships structured into five coherent clusters. While AI embeddings capture some of this categorical structure, they diverge substantially in global and local organization.

[4] ARI is a widely used metric for evaluating the similarity between two clustering assignments.

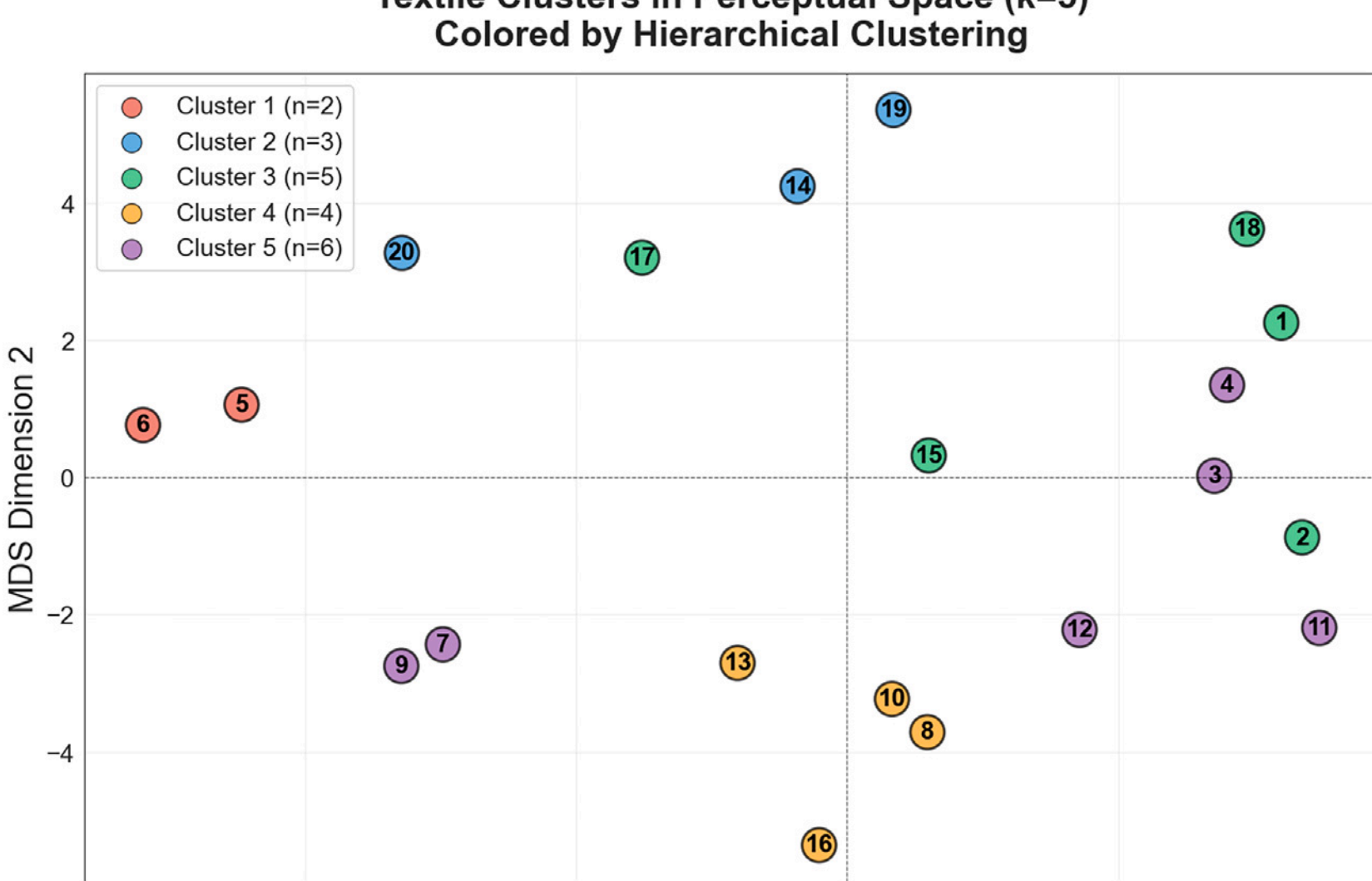

**Fig. C.12.** Two-dimensional MDS projection of perceptual similarity space based on human ratings. Colours indicate clusters ($k = 5$) derived using Ward's hierarchical clustering.

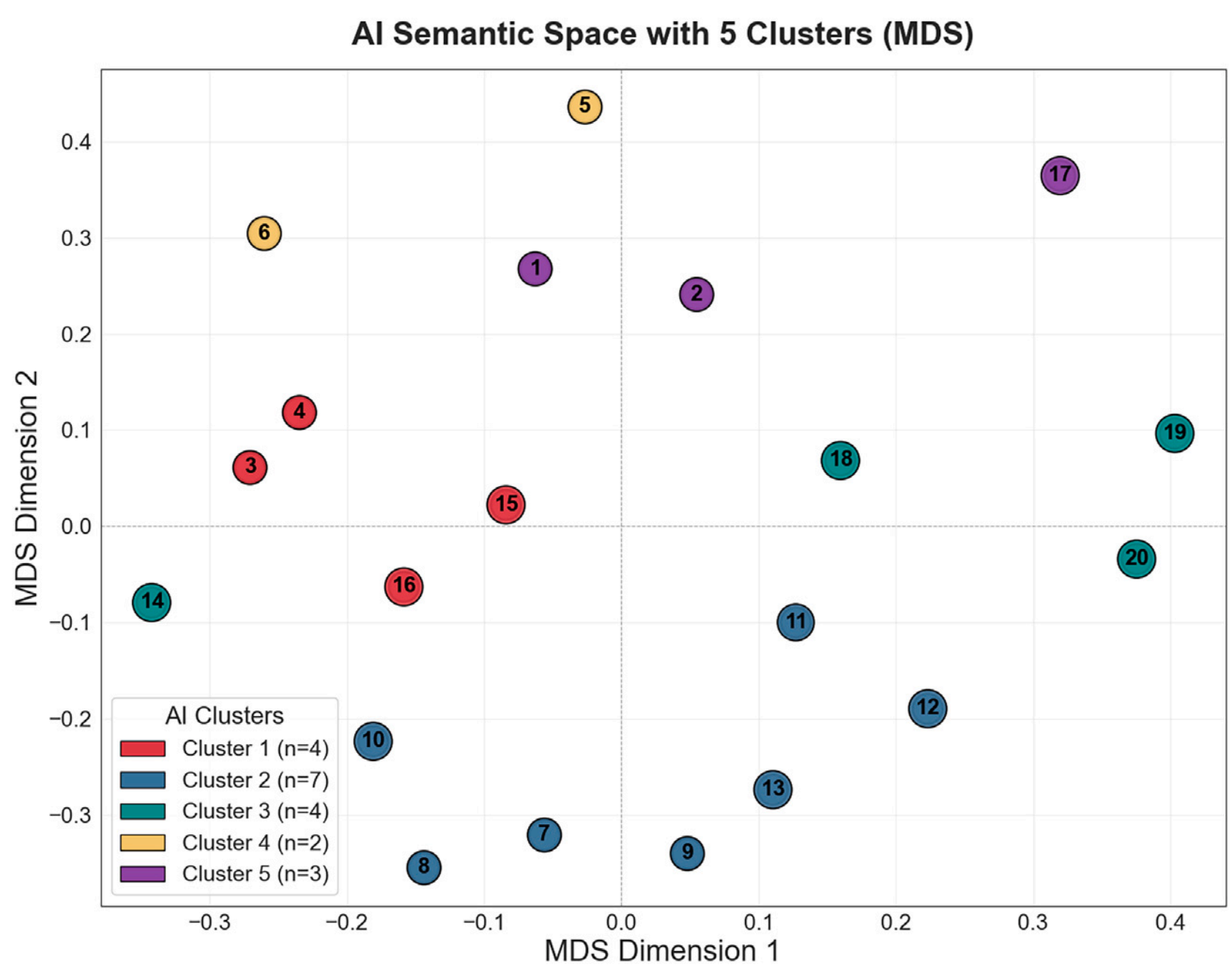

**Fig. C.13.** Two-dimensional MDS projection of the AI semantic similarity space derived from language model embeddings. Colours indicate clusters ($k = 5$) derived using Ward's hierarchical clustering.

**Table C.7**
Textile-level analysis of cluster alignment and task performance, ordered by success rate (descending). Alignment indicates whether a textile falls into the dominant AI cluster corresponding to its human perceptual cluster.

| Textile ID | Textile name | Human cluster | AI cluster | Alignment | Success rate (%) |
|---|---|---|---|---|---|
| 8 | slik_santin | 4 | 2 | ✓ | 100.0 |
| 6 | Melton_wool | 1 | 4 | ✓ | 50.0 |
| 10 | viscose_batiste | 4 | 2 | ✓ | 50.0 |
| 15 | nylon_taffeta | 3 | 1 | ✗ | 50.0 |
| 17 | cattle_leather | 3 | 5 | ✓ | 50.0 |
| 3 | linen_plain | 5 | 1 | ✗ | 50.0 |
| 4 | linen_twill | 5 | 1 | ✗ | 25.0 |
| 7 | slik_crepon | 5 | 2 | ✓ | 25.0 |
| 11 | cupro_plain | 5 | 2 | ✓ | 25.0 |
| 12 | cupro_charmeuse | 5 | 2 | ✓ | 25.0 |
| 1 | cotton_denim | 3 | 5 | ✓ | 0.0 |
| 2 | cotton_corduroy | 3 | 5 | ✓ | 0.0 |
| 5 | wool_houndstooth | 1 | 4 | ✓ | 0.0 |
| 9 | viscose_satin_stripe_organza | 5 | 2 | ✓ | 0.0 |
| 13 | polyester_organza | 4 | 2 | ✓ | 0.0 |
| 14 | polyester_three-layer composite knitting | 2 | 3 | ✓ | 0.0 |
| 16 | nylon_ultra-light_and_thin_fabric | 4 | 1 | ✗ | 0.0 |
| 18 | pp_non-woven-fabric | 3 | 3 | ✗ | 0.0 |
| 19 | PU_faux_leather | 2 | 3 | ✓ | 0.0 |
| 20 | polyester_faux_rabbit_fur | 2 | 3 | ✓ | 0.0 |

## Data availability

The data that has been used is confidential.